%% file: main.tex
\documentclass[11pt]{article}

\usepackage[final]{acl}

\usepackage{times}
\usepackage{latexsym}

\usepackage[T1]{fontenc}

\usepackage[utf8]{inputenc}

\usepackage{microtype}

\usepackage{inconsolata}

\usepackage{graphicx}

\title{Exploring \dtext Homology for Sentence Similarity}

\author{Marius Huber \quad Juri Opitz \\
  Department of Computational Linguistics \\
  University of Z{\"u}rich \\
  \texttt{\{marius.huber,jurialexander.opitz\}@uzh.ch}}

\usepackage{amssymb}
\usepackage{xspace}
\usepackage{calc}
\usepackage{cleveref}
\usepackage{enumitem}

\usepackage{adjustbox}
\usepackage{subcaption}

\newcommand{\RR}{\mathbb{R}}

\newcommand{\drtext}{Dowker--Rips\xspace}
\newcommand{\dtext}{Dowker\xspace}
\newcommand{\mean}{\mathrm{mean}}
\newcommand{\wght}{\mathrm{wght}}
\newcommand{\dow}{{\mathrm{Dow}}}

\graphicspath{{figures/}}

\hypersetup{
  pdftitle={Exploring Dowker Homology for Sentence Similarity},
  pdfauthor={Marius Huber and Juri Opitz}
}

\begin{document}
\maketitle
\begin{abstract}
  \input{abstract.tex}
\end{abstract}


\section{Introduction}\label{sec:introduction}


\dtext homology (DH;~\citealp{dowker_homology_groups_of_relations,chowdhury_memoli_functorial_dowker_theorem}) is a tool from topological data analysis, an emerging data analysis framework drawing on ideas from topology, metric geometry, and related areas of mathematics.
Among other things, DH may be used to capture the relative position of two point clouds living in a common space that is endowed with a distance function (typically \(\RR^{n}\) with Euclidean or cosine distance).

In this paper, we study a novel application of DH to NLP by applying it to Siamese sentence transformer models~\cite{reimers-gurevych-2019-sentence}.
Specifically, we ask the following.

\begin{enumerate}[label=(\arabic*), ref=(\arabic*), itemsep=0em, topsep=2pt]
  \item\label{question:distinguishing} Can DH distinguish similar and dissimilar sentence pairs?
  \item\label{question:information} Can DH be used to extract sentence similarity from token embeddings?
\end{enumerate}

On the positive side, we find that DH can elucidate sentence similarity data and models, both quantitatively and visually.
Findings regarding the second question are mixed: we define single-number summaries extracted from DH that capture ground truth sentence similarity reasonably well, but not more so than established methods.


\section{Background on \dtext Homology}\label{sec:background_dowker_rips_homology}

This section provides a high-level review of DH; for details, we refer the reader to Appendix~\ref{appendix:dowker_rips_homology}.
We limit ourselves to reviewing zero-dimensional DH, as that is the version of DH used in our experiments.

We consider a pair of point clouds \(X_{1}\) and \(X_{2}\) living in a common space that is endowed with a distance function \(d\).
We say that a point in \(X_{1}\) or \(X_{2}\) is \emph{\(r\)-close} to another if the distance between the two points is at most \(r\) (as measured by \(d\)).
Given a scale \(r\geq 0\), one constructs a graph whose vertices are those points in \(X_{1}\) that have a neighbor in \(X_{2}\) that is \(r\)-close, and where two points in \(X_{1}\) are connected by an edge whenever there exists a point in \(X_{2}\) that is \(r\)-close to \emph{both} of them.
In particular, the graph is constructed on elements of \(X_{1}\) only, while \(X_{2}\) merely dictates the connectivity of the graph.
To obtain DH of \(X_{1}\) relative to \(X_{2}\), one initially sets the scale to \(r=0\), and gradually increases it to theoretical infinity.
Along the way, one keeps track of the number of connected components constituting the aforementioned graph: any such component ``is born'' at a certain scale, and ``dies'' at some later scale as it is absorbed by---that is, connected via an edge to---an existing component.
This yields a sequence \((b_{1},d_{1}),\dots,(b_{k},d_{k})\), where each pair contains the scales at which the birth and death, respectively, of a component happen.
This sequence is recorded in a \emph{persistence diagram}, which is obtained by placing a point at coordinates \((b_{i},d_{i})\) in \(\RR^{2}\) for all \(i=1,\dots,k\).\footnote{
  A persistence diagram is usually endowed with the diagonal line where birth values equal death values.
  The vertical distance of the \(i\)-th point to the diagonal equals \(d_{i}-b_{i}\), and can thus be thought of as the ``lifetime'' of the corresponding connected component, \(i=1,\dots,k\).
}
A key property of DH is that it is unaffected when swapping the roles of \(X_{1}\) and \(X_{2}\) in the above process, making the resulting persistence diagram an ``undirected'' quantity~\cite{dowker_homology_groups_of_relations,chowdhury_memoli_functorial_dowker_theorem}.

\begin{figure}[t]
  \centering
  \includegraphics[width=\linewidth]{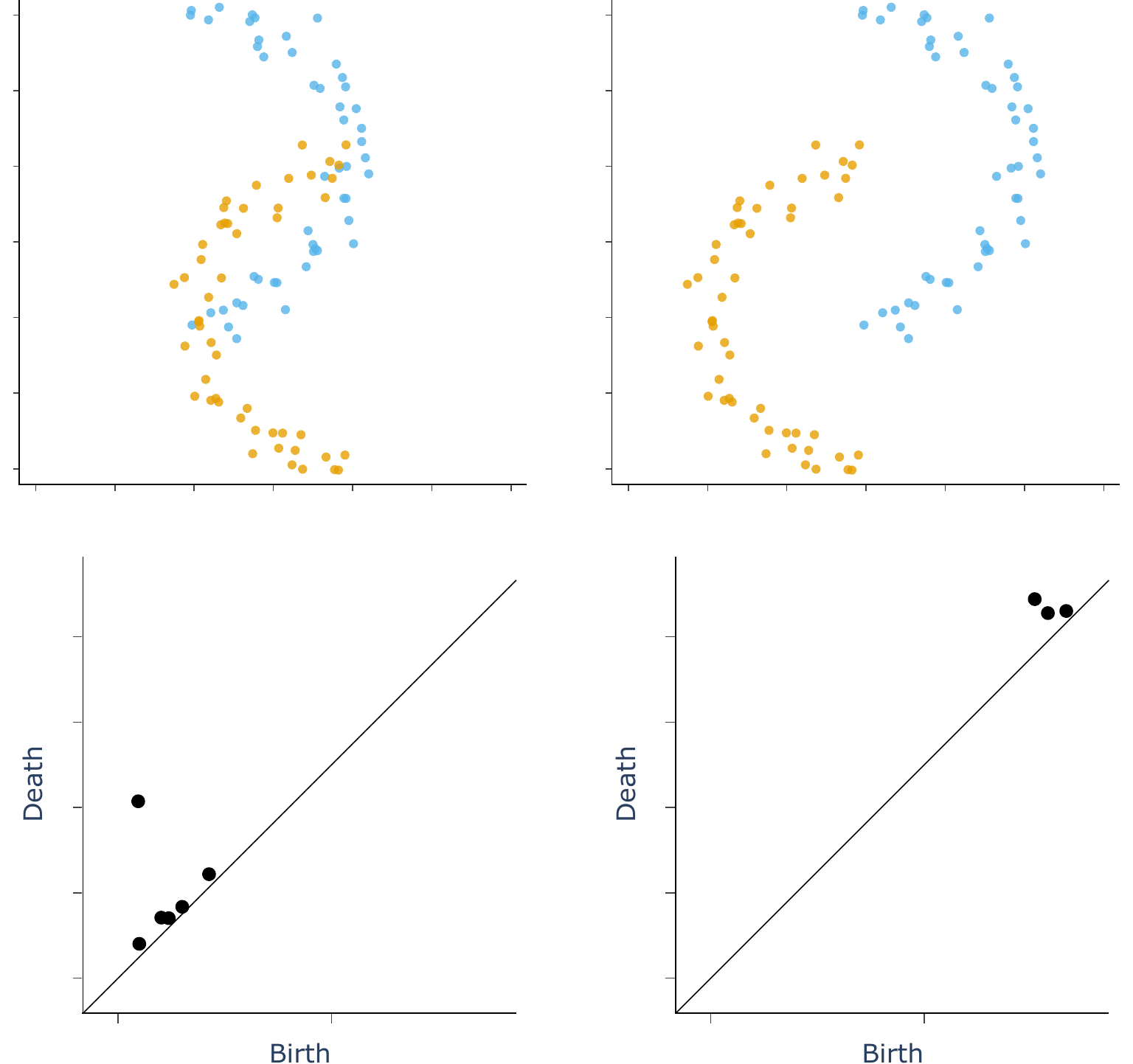}
  \caption{
    Two pairs of point clouds of varying degree of spatial separation (top) and their corresponding \dtext persistence diagrams (bottom).
    The more the point clouds are separated, the larger the birth values in the persistence diagrams.
  }
  \label{fig:dowker_rips_distance_example}
\end{figure}

If the point clouds \(X_{1}\) and \(X_{2}\) are spatially well separated, vertices and edges appear only at large scales.
If \(X_{1}\) and \(X_{2}\) nearly coincide, those appear at small scales already; indeed, if \(X_{1}=X_{2}\), all birth values equal zero.
The spatial proximity (resp. separation) of \(X_{1}\) and \(X_{2}\) is thus reflected in the points of the \dtext persistence diagram having small (resp. large) birth values; see Figure~\ref{fig:dowker_rips_distance_example}.


\section{\dtext Homology for  Similarity}\label{sec:dowker_homology_for_similarity}

\begin{figure}[t]
  \centering
  \includegraphics[width=\columnwidth]{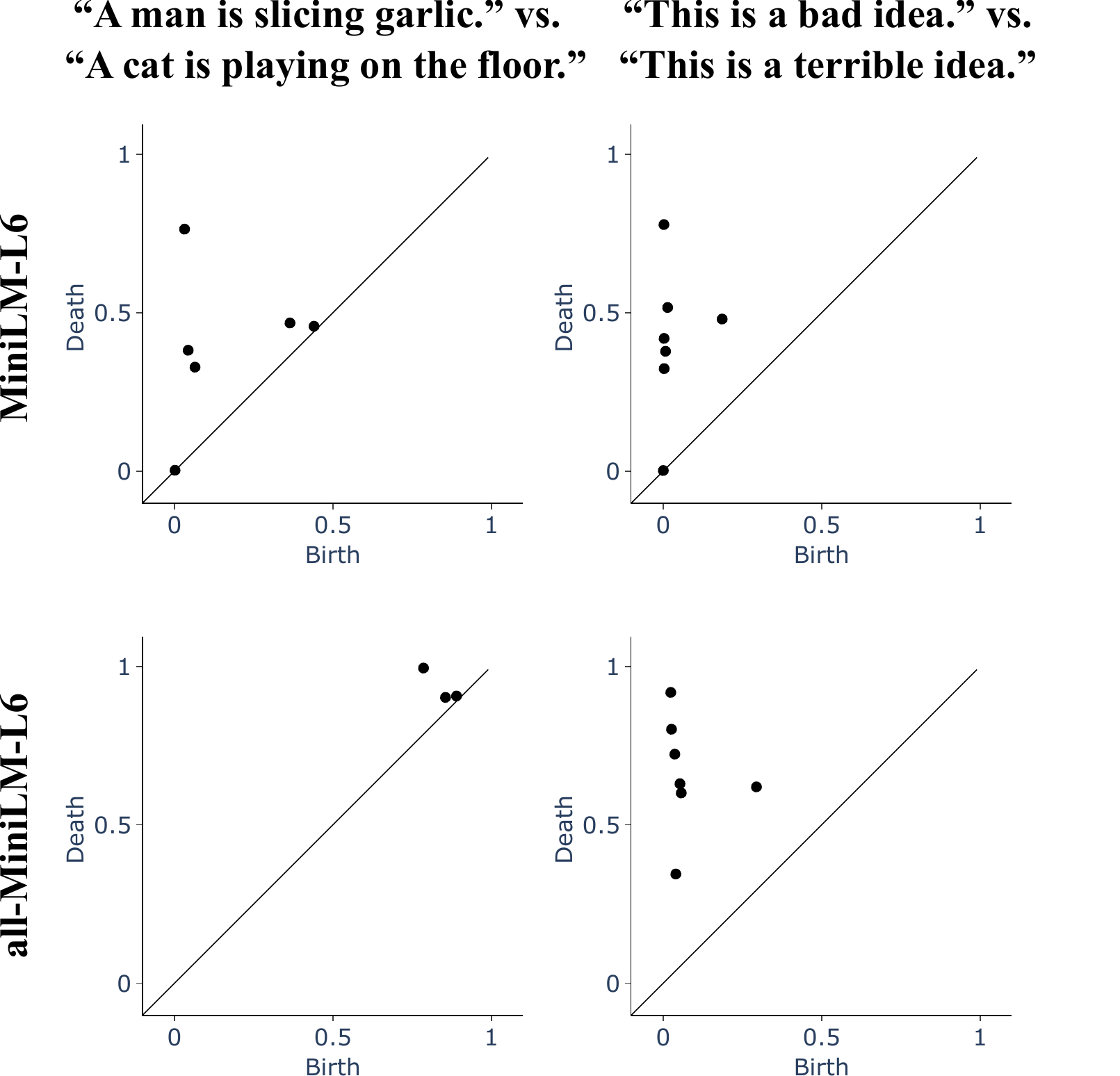}
  \caption{
    \dtext persistence diagrams capturing sentence similarity of two randomly selected sentence pairs from the STS-B dataset with low (left column) and high (right column) similarity, computed from embeddings produced using MiniLM-L6 (top row) and all-MiniLM-L6 (bottom row).
    Higher sentence similarity is reflected in smaller birth values in the diagrams, and the fact that MiniLM-L6 is not fine-tuned for sentence similarity is reflected in the top two diagrams being more similar to each other than the bottom two diagrams.
    Moreover, fine-tuning MiniLM-L6 for sentence similarity substantially alters the persistence diagram corresponding to the dissimilar sentence pair, while leaving the other diagram largely unaffected.
  }
  \label{fig:persistence_grid}
\end{figure}

To apply DH to sentence similarity, we treat the token embeddings stemming from a pair of sentences as a pair of point clouds in the latent space of a transformer model, and compute the corresponding \dtext persistence diagram.
If the two sentences are similar (resp. dissimilar), we expect the corresponding point clouds to be colocalized (resp. separated) and, as discussed in Section~\ref{sec:background_dowker_rips_homology}, we expect the resulting \dtext persistence diagram to have predominantly small (resp. large) birth values.

We point out that our method is similar in spirit to those used in~\citet{barannikov_representation_topology_divergence} and~\citet{michel_geometry_of_word_embeddings}.
Indeed, both methods also draw on ideas from topological data analysis to assess the (dis-)similarity of two point clouds.
The former method, however, requires the two point clouds to be of equal sizes, while the latter takes into account only the topological features of the two point clouds individually and does not capture, for instance, their spatial separation.

\begin{figure}[t]
  \centering
  \includegraphics[width=\linewidth]{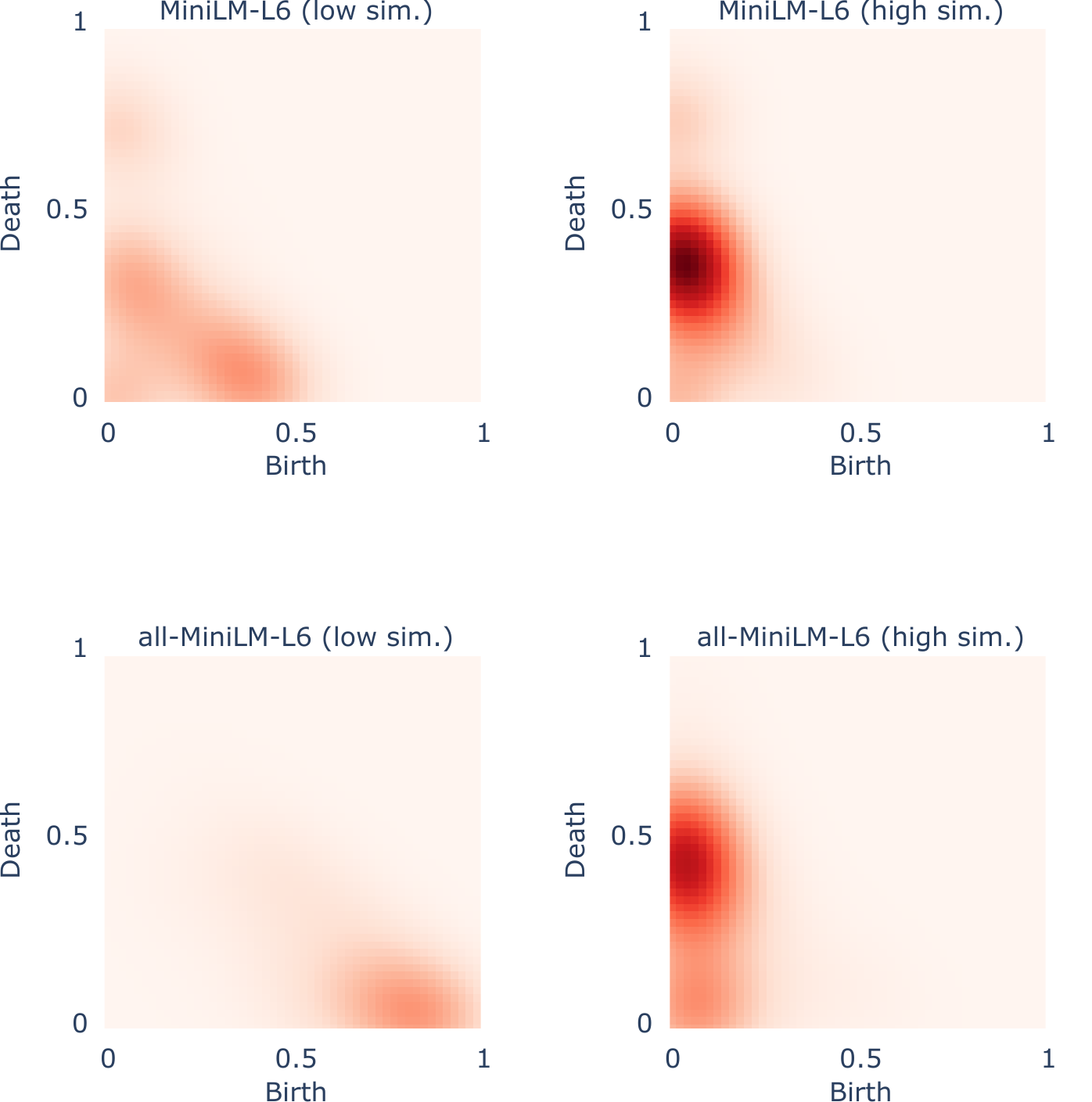}
  \caption{
    Mean persistence images capturing sentence similarity of all sentence pairs from the STS-B dataset with low (left column) and high (right column) similarity, computed from embeddings produced using MiniLM-L6 (top row) and all-MiniLM-L6 (bottom row).
    These persistence images can be thought of as ``heat maps'' that capture the distribution of points across multiple persistence diagrams (see Appendix~\ref{appendix:vectorizing_persistence_diagrams} for details).
    The images indicate that fine-tuning MiniLM-L6 for sentence similarity leads to larger birth values in DH of dissimilar sentence pairs, thus generalizing the findings from Figure~\ref{fig:persistence_grid} across the entire STS-B dataset.
    Here, the sentence pairs with low (resp. high) similarity are those whose ground truth similarity in STS-B differs by at most \(0.1\) from the minimum (resp. maximum) possible similarity value.
    For the analogous plots for the other models used, see Appendix~\ref{appendix:mean_persistence_images}.
  }
  \label{fig:mean_persistence_images_minilm_l6}
\end{figure}

\paragraph{Example.} Figure~\ref{fig:persistence_grid} illustrates that our expectation holds for two sentence pairs randomly selected from the STS Benchmark (STS-B;~\citealp{cer-etal-2017-semeval}) when using the fine-tuned all-MiniLM-L6 model, but not when using the non-fine-tuned MiniLM-L6 model.
Fine-tuning a transformer model for sentence similarity seems to strongly affect how dissimilar sentences are represented (as seen through the lens of DH), but less so for similar sentences, where the persistence diagrams stemming from the non-fine-tuned and the fine-tuned model are basically indistinguishable.
Indeed, the finding of Figure~\ref{fig:persistence_grid} generalizes across all similar and dissimilar pairs from the STS-B dataset, as illustrated in Figure~\ref{fig:mean_persistence_images_minilm_l6}.

\paragraph{Experiment outline.} Motivated by this example, we conduct the following two experiments for each layer of every model from a diverse set of transformer models fine-tuned for sentence similarity as well as their underlying, non-fine-tuned base models (see Table~\ref{table:models} in Appendix \ref{appendix:models_used} for a list of models used).
We use sentence pairs from the STS-B dataset, which contains sentence pairs annotated by human raters with semantic similarity scores ranging from \(0\) to \(5\).

\begin{enumerate}[itemsep=0em, topsep=2pt]
  \item \textbf{Layer regression experiment:} We investigate to what extent standard token-level measures as well as \dtext persistence diagrams contain information on sentence similarity.
        We do so by fitting a linear regression model that predicts ground truth sentence similarities from these measures.
  \item \textbf{Layer correlation experiment:} In order to circumvent the need for a regression trained on \dtext persistence diagrams, and to provide measures of sentence similarity that are more readily applicable than the raw persistence diagrams, we define single-number summaries extracted from DH.
        These summaries are defined so that we expect them to capture sentence similarity directly, and we assess how well they correlate with ground truth sentence similarities.
\end{enumerate}

We now describe the sentence similarity methods used in our experiments.


\subsection{Baseline Similarity Methods}\label{sec:baseline_methods}

We use BOS-, EOS-, max- and mean-pooling as single-vector sentence representations.
The similarity of a sentence pair is then computed as the cosine similarity between the pooled representations of the sentences.


\subsection{\dtext Similarity}\label{sec:dowker_similarity}

We define four versions of \emph{\dtext similarity} (DS), one for each of four different kinds of aggregation used along the way.
To compute DS for a sentence pair \((s_{1},s_{2})\), we consider the token embeddings of \(s_{1}\) and \(s_{2}\) as point clouds \(X_{1}\) and \(X_{2}\), respectively, and compute DH of \(X_{1}\) relative to \(X_{2}\), as explained in Section~\ref{sec:background_dowker_rips_homology}.
We extract single-number summaries from the resulting persistence diagram by computing the minimum, maximum and average of all birth values of points in the diagram, as well as the weighted average birth value, where birth values are weighted by the lifetime of the corresponding point.
We regard each of these as a notion of distance between \(X_{1}\) and \(X_{2}\), since larger birth values indicate a larger degree of spatial separation of \(X_{1}\) and \(X_{2}\) (see Section~\ref{sec:background_dowker_rips_homology}), and we denote these distances by \(d_{\dow}^{\min}\), \(d_{\dow}^{\max}\), \(d_{\dow}^{\mean}\) and \(d_{\dow}^{\wght}\), respectively.
Finally, we define the DSs by setting \(s_{\dow}^{\bullet}(s_{1},s_{2})=e^{-d_{\dow}^{\bullet}(X_{1},X_{2})}\) for \(\bullet\in\left\{\min,\max,\mean,\wght\right\}\).
All versions of DS thus attain values between \(0\) and \(1\).
Moreover, if \(s_{1}=s_{2}\), then \(X_{1}=X_{2}\), and all birth values in DH equal zero, so that all four versions of \dtext distance vanish, and, correspondingly, all four versions of DS yield a similarity value of \(1\) for identical sentences.


\section{Experiments and Results}\label{sec:experiments_and_results}

We now provide detailed descriptions of the two experiments conducted, and then briefly discuss some limitations and design choices.


\subsection{Layer Regression Experiment}\label{sec:layer_regression_experiment}

\begin{figure*}[t]
  \centering
  \includegraphics[width=0.49\linewidth]{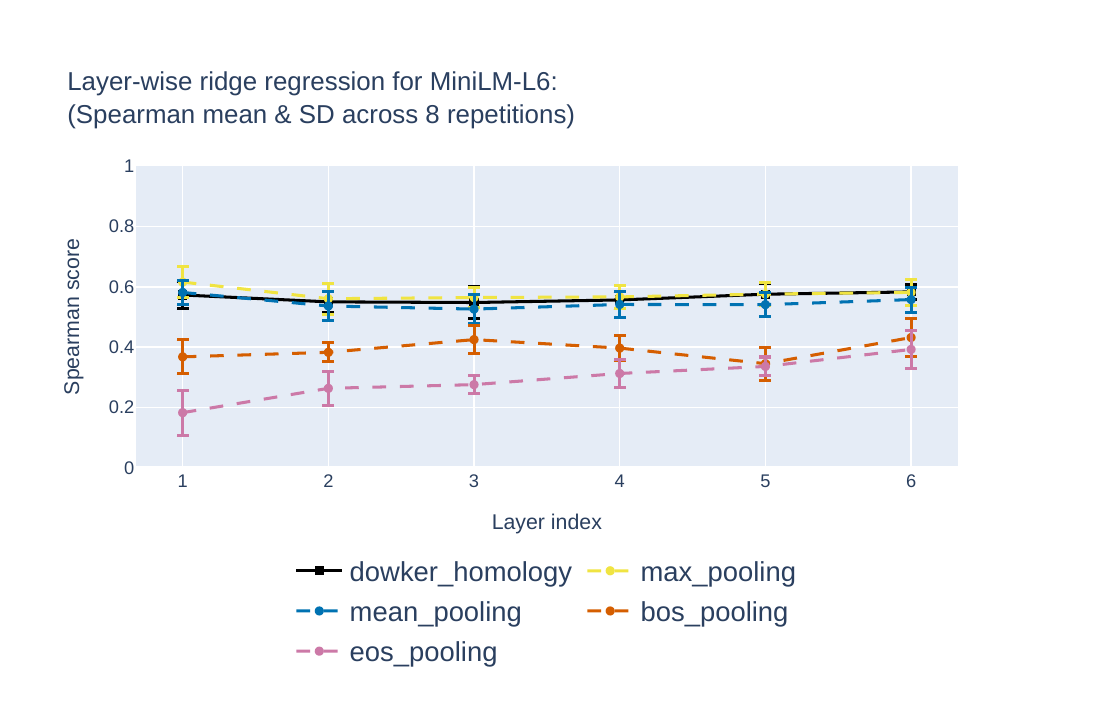}
  \hfill
  \includegraphics[width=0.49\linewidth]{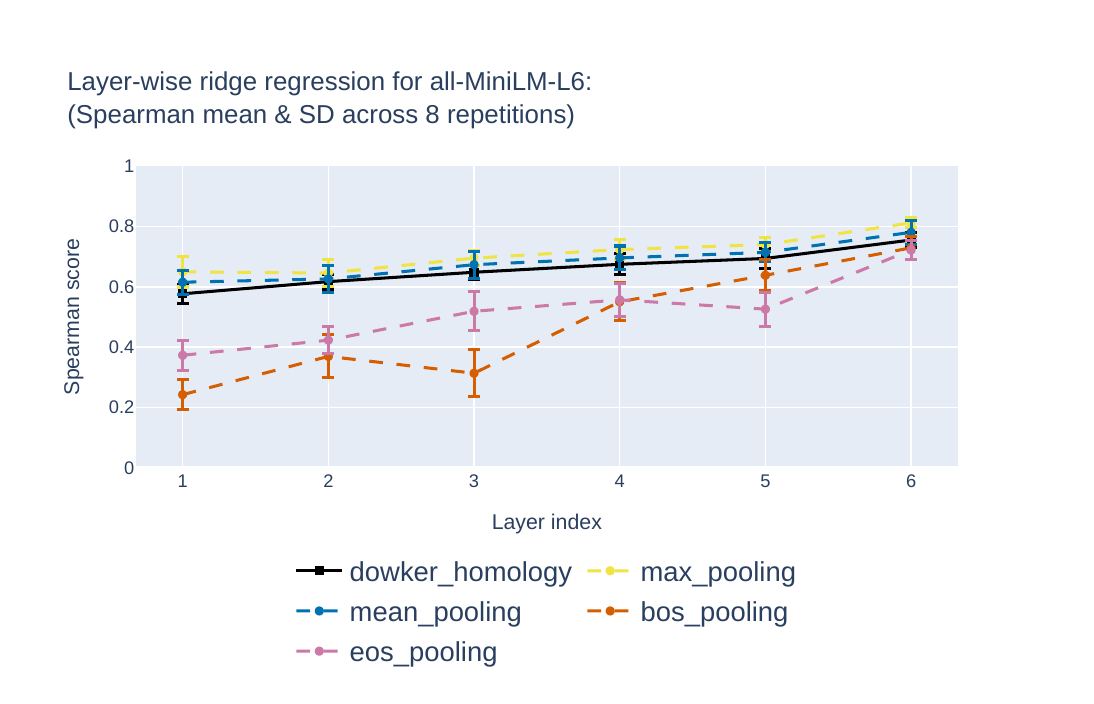}
  \caption{
    Two results of the layer regression experiment.
    Bars indicate standard deviation over eight runs.
  }
  \label{fig:layer_regression_minilm_l6}
\end{figure*}

\paragraph{Setup.} We sample \(512\) and \(256\) sentence pairs from the train and test splits, respectively, of STS-B.
As baseline features, we use the vectors \((u_{1}, u_{2}, \left|u_{1}-u_{2}\right|)\), where \(u_{1}\) and \(u_{2}\) are the pooled representations of a sentence pair \((s_{1},s_{2})\), obtained via the four pooling methods described in Section~\ref{sec:baseline_methods} (this is motivated by~\citet[Section~6]{reimers-gurevych-2019-sentence}; here, \(\left|\,\cdot\,\right|\) denotes the entry-wise absolute value).
For DH, we compute the persistence diagram of each sentence pair from the corresponding token embeddings (as described in Section~\ref{sec:dowker_similarity}) using cosine distance, and vectorize it into a persistence image of resolution \(25\times 25\), resulting in a vector of length \(625\) (see Appendix~\ref{appendix:vectorizing_persistence_diagrams} for details on persistence images).
We then train a ridge regression on these features and evaluate it using Spearman's rank correlation between the predicted and the ground-truth similarities; the choice of ridge regression is motivated by the fact that persistence images produce high-dimensional, sparse vectors with correlated features, making ordinary least squares estimates potentially unstable (see Appendix~\ref{appendix:hyperparameter_selection} for selection of hyperparameters).
This procedure is run for each hidden layer of each model.

\paragraph{Results.} We find that across all models (fine-tuned or not), the regression trained on the \dtext persistence diagrams never outperforms the one trained on the best performing baseline similarity methods.
The regression trained on the \dtext persistence diagrams always wins at least third place (as measured by Spearman correlation averaged across all layers of a model), consistently beating those trained on BOS- and EOS-pooling, but coming short of outperforming those trained on max- and mean-pooling.
Interestingly, max-pooling yields the best features for \(9/10\) models despite the fact that the sentence transformers used were fine-tuned using mean-pooling.
Finally, we observe that in the fine-tuned models---in contrast to the non-fine tuned ones---all similarity methods tend to yield higher Spearman correlations on later layers, which is somewhat intuitive as the final layer is where fine-tuning is directly applied.
As an example, Figure~\ref{fig:layer_regression_minilm_l6} shows the Spearman correlations of all methods at each layer of MiniLM-L6 (not fine-tuned) and all-MiniLM-L6 (fine-tuned); the former is one of the two cases in which the features derived from DH rank second, outperforming those derived from mean-pooling.
Appendix~\ref{appendix:layer_regression_results} shows analogous plots for the other model pairs.


\subsection{Layer Correlation Experiment}\label{sec:layer_correlation_experiment}

\begin{figure*}[t]
  \centering
  \includegraphics[width=0.49\linewidth]{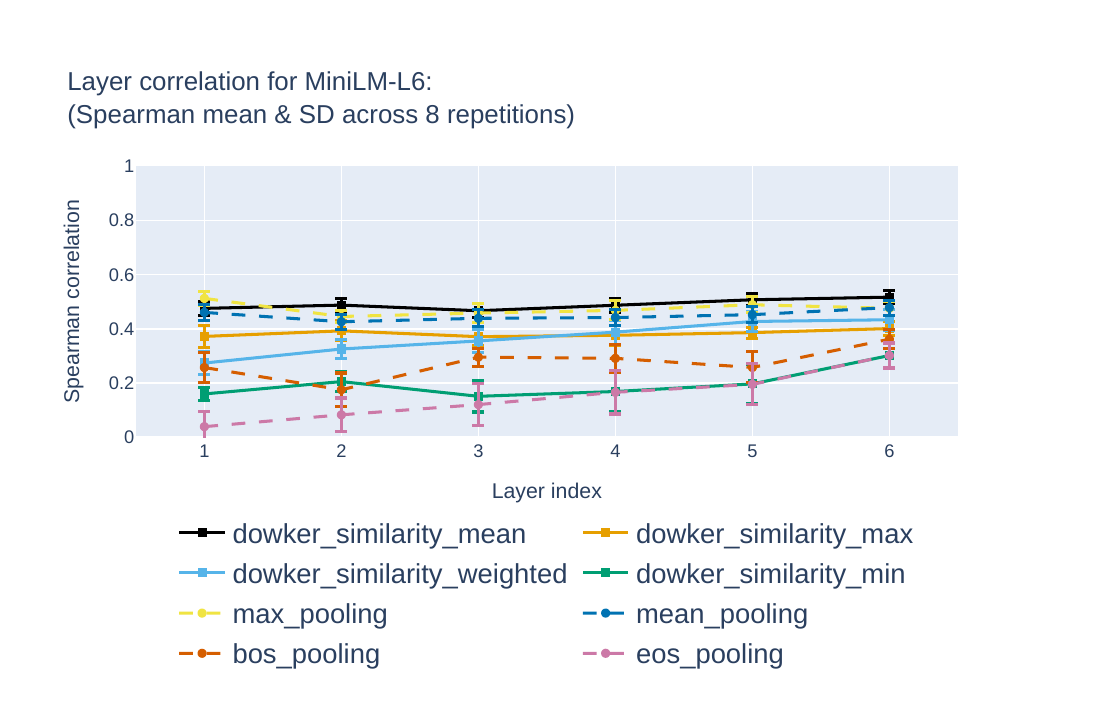}
  \hfill
  \includegraphics[width=0.49\linewidth]{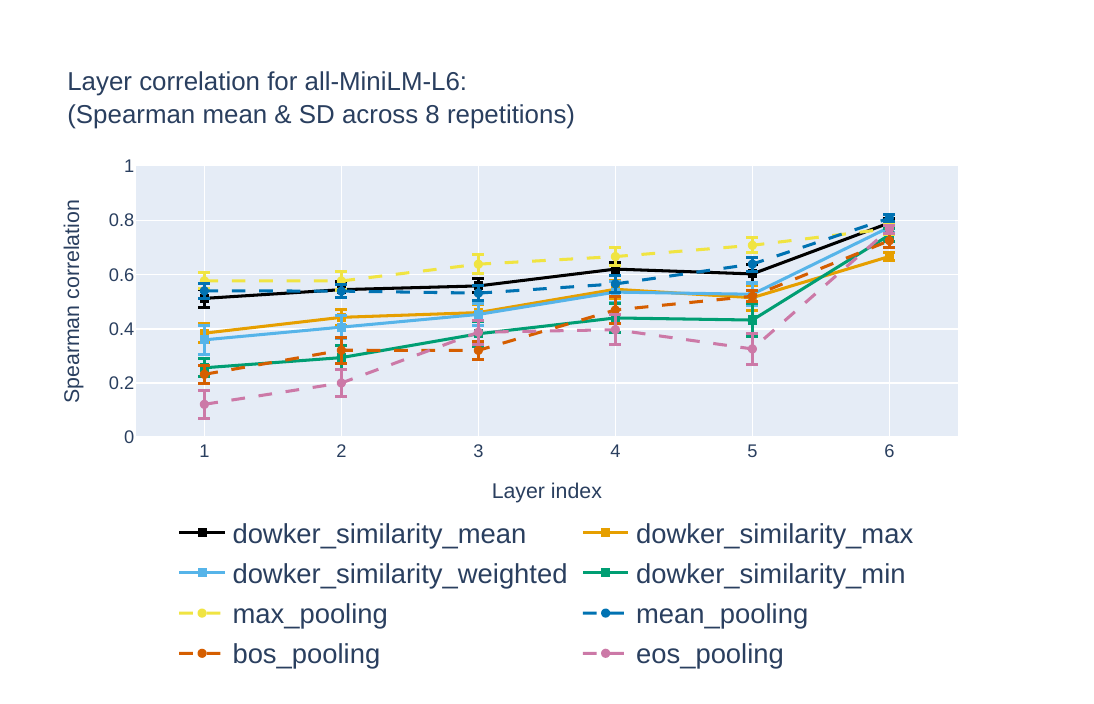}
  \caption{
    Two results of the layer correlation experiment.
    Versions of \dtext similarity are indicated by the suffixes.
  }
  \label{fig:layer_correlation_minilm_l6}
\end{figure*}

\paragraph{Setup.} We sample \(256\) sentence pairs from the test split of STS-B, and compute similarities for each hidden layer of each model, using the four baseline similarity methods and the four versions of DS from Sections~\ref{sec:baseline_methods} and~\ref{sec:dowker_similarity}, respectively.
For each method, model and layer, we evaluate the performance via Spearman correlation between the predicted and the ground-truth similarities.

\paragraph{Results.} As before, we assess the goodness of the baseline methods and the four versions of DS by the Spearman correlations averaged across all layers of a model.
We find that (at least one version of) DS comes in third behind the max- and mean-pooling methods for \(5/10\) models, and wins first and second place for \(2/10\) models each.
Again, we find that max-pooling outperforms mean-pooling in terms of average Spearman correlation.
Mean-pooling, however, unsurprisingly yields the highest maximal Spearman correlation for the sentence transformer models, and this is always the case at the final model layer, sometimes after a sheer increase of performance in the last few layers.
Finally, we find that \(s_{\dow}^{\mean}\) outperforms the other versions of DS for \(9/10\) models.
As an example, Figure~\ref{fig:layer_correlation_minilm_l6} shows the Spearman correlations of all methods at each layer of MiniLM-L6 and all-MiniLM-L6; the former is one of two cases where \(s_{\dow}^{\mean}\) outperforms all other methods (the other such instance being MiniLM-L12).
We refer to Appendix~\ref{appendix:layer_correlation_results} for plots of the remaining results.


\subsection{Experiment Limitations}\label{sec:experiment_limitations}

\paragraph{Dataset and Sample Sizes.} In both experiments, we confined ourselves to working with just the STS-B dataset and the relatively modest respective sample sizes due to computational constraints.

\paragraph{Computation of DH Feature Vectors.} While DH may be computed using any metric, we restricted ourselves to using cosine distance in our experiments.
We experimented using Euclidean distance, but cosine distance consistently yielded better results.
For the layer regression experiment, the resolution of \(25\times 25\) used for vectorization of persistence diagrams was chosen as an empirical compromise between preserving performance and limiting the dimensionality of the resulting feature vectors.
Finally, we restricted ourselves to working with zero-dimensional DH because we suspected higher-dimensional DH to capture information that goes beyond spatial separation of point clouds.
Indeed, preliminary experiments showed that the point clouds consisting of token embeddings rarely exhibited any higher-dimensional DH at all, and including higher-dimensional DH as features in the layer regression experiment had virtually no effect on the results.

\paragraph{Evaluation metric.} In both experiments, we confined ourselves to evaluating the agreement of DSs with ground truth similarities via Spearman correlation since DSs are obtained by applying the monotonic transformation \(d\mapsto e^{-d}\) to the uncalibrated distance measures \(d_{\dow}^{\min}\), \(d_{\dow}^{\max}\), \(d_{\dow}^{\mean}\) and \(d_{\dow}^{\wght}\).
We hence do not expect any linear relationship between DSs and ground truth similarities as captured, for instance, by Pearson correlation.


\section{Discussion}\label{sec:discussion}

Our experiments show that \dtext persistence diagrams reflect sentence similarity (cf. Question~\ref{question:distinguishing}) and, moreover, that they can provide visual insights on NLP models, such as the effect of fine-tuning on the representation of similar and dissimilar sentence pairs (cf. the example in Section~\ref{sec:dowker_homology_for_similarity}).
Moreover, while all versions of DS capture sentence similarity to some extent, the version that averages birth values is by far the best-performing one, as witnessed by the fact that it correlates with ground truth similarities essentially as well as a trained linear regression.
Even this version of DS, however, does not consistently outperform established methods, answering Question~\ref{question:information} with a ``yes, but''.
In summary, our findings provide insights on the usefulness and interpretability of DH, and justify further research on connections between topology and sentence similarity, such as exploration of further single-number summaries extracted from DH.



\bibliography{references}


\appendix

\renewcommand{\thefigure}{A.\arabic{figure}}
\setcounter{figure}{0}
\renewcommand{\thetable}{A.\arabic{figure}}
\setcounter{table}{0}

\section{Appendix}\label{sec:appendix}


\subsection{\dtext Homology}\label{appendix:dowker_rips_homology}

DH, introduced by~\citet{dowker_homology_groups_of_relations} and promoted to the setting of persistent homology by~\citet{chowdhury_memoli_functorial_dowker_theorem}, is a tool from topological data analysis that captures the topology of two sets \(X\) and \(Y\) and a relation \(R\subseteq X\times Y\) between them.
Here, we review DH as used in the main text of the article.
We point out that the implementation of DH used in our experiments is that of \drtext homology provided by~\citet{huber_schnider_flagifying_dowker_complex}.
While \drtext homology is, in general, merely an approximate, computationally cheaper variant of DH, we use that implementation since all our experiments rely only on zero-dimensional DH, in which case \dtext and \drtext homology coincide.
Correspondingly, the following exposition is restricted to zero-dimensional DH.

In the case where \(X\) and \(Y\) are subsets of some metric space \((Z,d)\), and given some scale \(r\geq 0\), one may define a relation \(R_{r}\) by declaring that \((x,y)\in R_{r}\) iff \(d(x,y)\leq r\), for \(x\in X\) and \(y\in Y\).
The relation \(R_{r}\) may be translated into the graph \(\Gamma_{r}=(V_{r},E_{r})\), defined as having vertex set \[
  V_{r}=\left\{x\in X\mid\exists y\in Y\colon d(x,y)\leq r\right\}\subseteq X,
\] and where two vertices \(x_{1},x_{2}\in X\) are connected via an edge whenever there exists \(y\in Y\) such that \(d(x_{1},y)\leq r\) and \(d(x_{2},y)\leq r\).

As is the case for any graph, the graph \(\Gamma_{r}\) consists of some---possibly zero---connected components.
Indeed, if \(r=0\), \(\Gamma_{r}=\varnothing\) (provided that \(X\) and \(Y\) have no common elements).
As one increases the scale \(r\), vertices and edges appear in \(\Gamma_{r}\).
Correspondingly, the number of connected components of \(\Gamma_{r}\) can increase or decrease as \(r\) increases: a new vertex may appear when \(r\) becomes large enough for that vertex to ``see'' an element of \(Y\), or two existing components of \(\Gamma_{r}\) may merge into one when \(r\) becomes large enough for two vertices initially belonging to different connected components to be connected via an edge.
In other words, increasing the scale \(r\) from zero to (theoretical) infinity yields a sequence ``births'' and ``deaths'' of connected components.
DH is the information contained in this process, and may be recorded by a sequence \((b_{1},d_{1}),\dots,(b_{k},d_{k})\) of pairs of non-negative real numbers, where the \(i\)-th pair contains the value of the scale \(r\) at which the ``birth'' (\(b_{i}\)) and ``death'' (\(d_{i}\)), respectively, of the \(i\)-th connected component occur.

This sequence is usually recorded in a \emph{persistence diagram}, which, by definition, is the subset of \(\RR^{2}\) consisting of the pairs \((b_{1},d_{1}),\dots,(b_{k},d_{k})\) (see Figure~\ref{fig:dowker_rips_distance_example} in the main text for examples of persistence diagrams).
Persistence diagrams are usually endowed with the diagonal line characterized by those points where birth values equal death values; the vertical distance of the point \((b_{i},d_{i})\) in the persistence diagram indicates the lifetime (that is, \(d_{i}-b_{i}\)) of the \(i\)-th connected component.
As such, a persistence diagram provides an at-a-glance summary of the evolution of the graph \(\Gamma_{r}\) as \(r\) increases from zero to infinity, and---as explained in Section~\ref{sec:background_dowker_rips_homology}---captures the spatial separation of \(X\) and \(Y\) as measured by the metric \(d\) defined on the metric space \((Z,d)\).
We close this section by remarking that any \dtext persistence diagram contains precisely one point whose lifetime is infinite, that is, which dies at only scale \(r=\infty\).
Since this infinite value typically breaks many formulas and pipelines, one usually either discards this point from the diagram, or replaces the scale at which it dies by some finite number that is chosen according to some heuristic.
In our experiments, we experimented with both options, and found that simply discarding that point yielded better and more stable results.


\subsection{Vectorizing Persistence Diagrams}\label{appendix:vectorizing_persistence_diagrams}

\begin{figure*}[t]
  \centering
  \def\svgscale{1}
  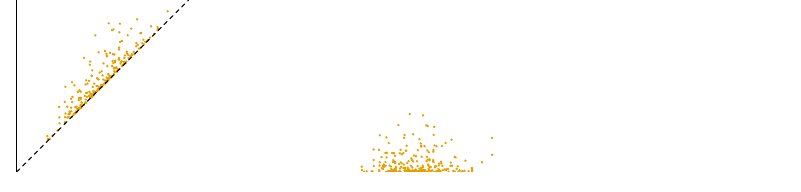
  \caption{
    Schematic showing the transformation of a persistence diagram in (birth, death)-coordinates (left) to (birth, death-birth)-coordinates (middle) by means of a shearing map, and the resulting persistence surface (right).
    Figure adapted from~\citet{huber_2026_fixation_sequences}.
  }
  \label{fig:persistence_image}
\end{figure*}

To make persistence diagrams usable in machine learning pipelines, one must transform them into vectors of a fixed length.
In our experiments, we use a method known as \emph{persistence image}~\cite{adams_persistence_images}.
A persistence image is obtained from a persistence diagram by first applying a shearing that maps the diagonal of the persistence diagram to a horizontal line.
The resulting diagram is then converted into a ``heat map'' by placing a bivariate Gaussian distribution with a given standard deviation (also referred to as \emph{bandwidth} in the present context) at each point of the diagram.
This yields a single-channel image commonly referred to as persistence surface; see Figure~\ref{fig:persistence_image} for an illustration of the process of obtaining a persistence surface from a persistence diagram.
Finally, the persistence image is obtained from the persistence surface by discretizing it into a pixel grid of a chosen \emph{resolution}.
The bandwidth and resolution used in the process are treated as hyperparameters in applications.


\subsection{Models used}\label{appendix:models_used}

\input{tables/table_models.tex}

See Table~\ref{table:models} for a list of models used in our experiments.

\subsection{Hyperparameter Selection}\label{appendix:hyperparameter_selection}

To train the ridge regressions used in the layer regression experiment described in Section~\ref{sec:layer_regression_experiment}, we used the implementation provided by scikit-learn~\cite{scikit-learn}.
In doing so, we used the SVD-solver, tuned the regularization parameter \(\alpha\) via grid search over the values \(10^{-3},10^{-2},\dots,10^{2}\), and left other parameters at their default values.
Persistence diagrams in these experiments were transformed into persistence images using the GUDHI library~\cite{gudhi:PersistenceRepresentations}.
We used default values for persistence image creation, apart from the bandwidth, which we tuned via grid search over the values \(10^{-3},10^{-2},\dots,10^{2}\).


\subsection{Results}\label{appendix:results}

Here we provide plots containing the results from the two experiments conducted, analogous to the plots in the main text.


\subsubsection{Mean Persistence Images for Dissimilar and Similar Sentence Pairs}\label{appendix:mean_persistence_images}

\begin{figure*}[t]
  \centering
  \begin{subfigure}{\columnwidth}
    \includegraphics[width=\columnwidth]{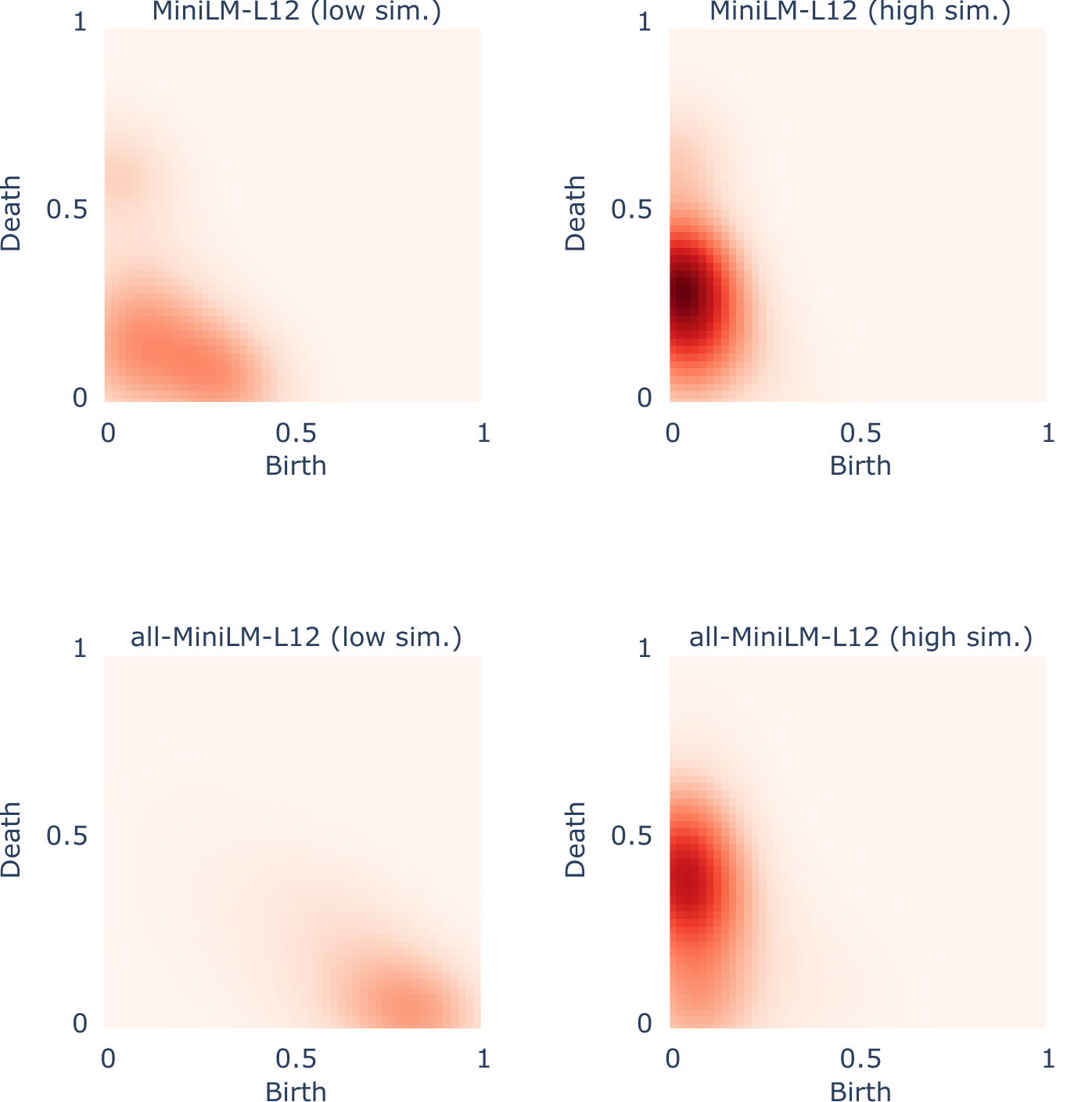}
    \captionsetup{width=0.9\columnwidth}
    \caption{
      Mean persistence images for the models MiniLM-L12-H384-uncased and all-MiniLM-L12-v2.
    }
    \vspace{1em}
    \label{fig:mean_persistence_images_minilm_l12}
  \end{subfigure}
  \begin{subfigure}{\columnwidth}
    \includegraphics[width=\columnwidth]{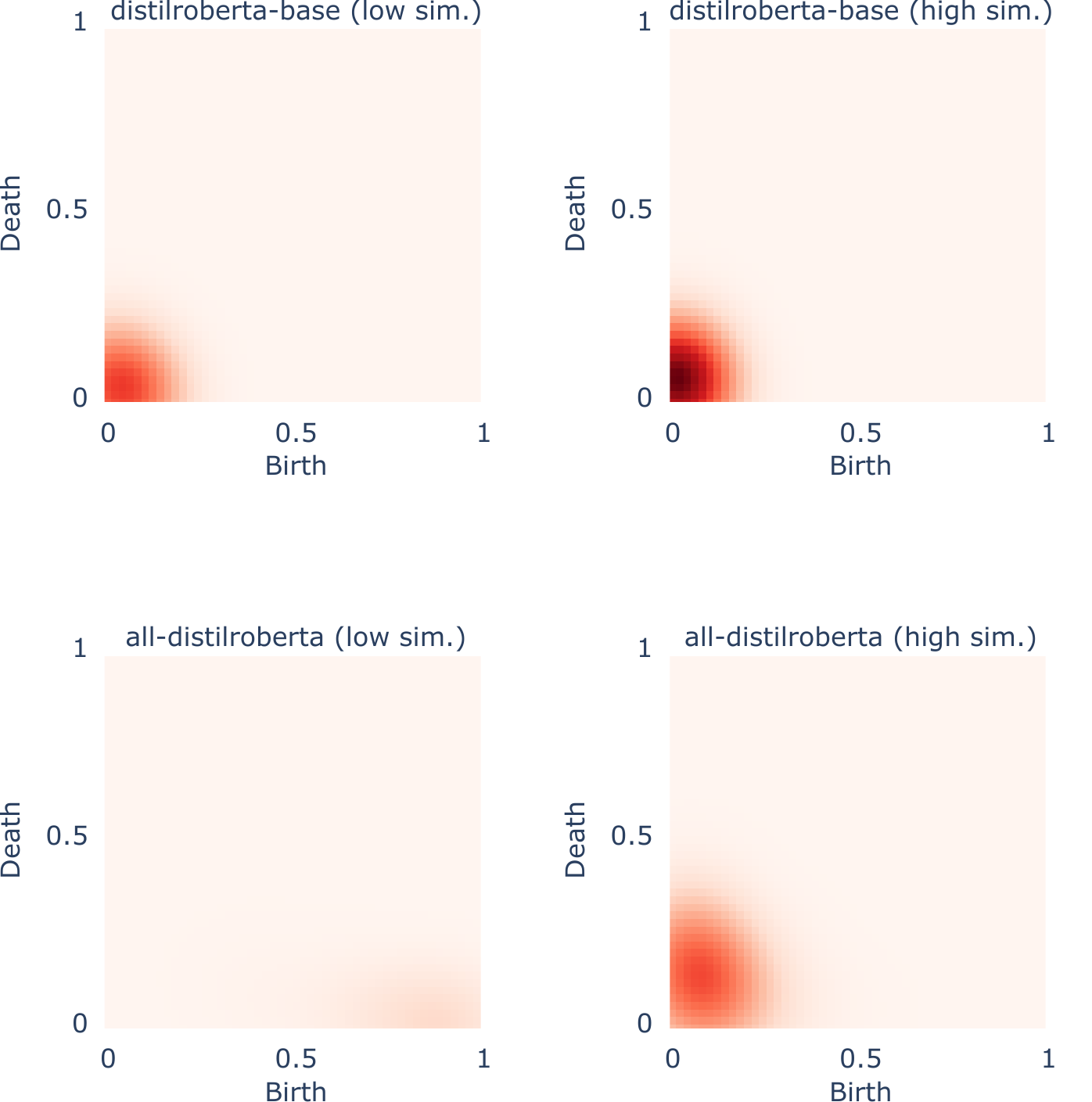}
    \captionsetup{width=0.9\columnwidth}
    \caption{
      Mean persistence images for the models distilroberta-base and all-distilroberta-v1.
    }
    \vspace{1em}
    \label{fig:mean_persistence_images_distilroberta}
  \end{subfigure}

  \begin{subfigure}{\columnwidth}
    \includegraphics[width=\columnwidth]{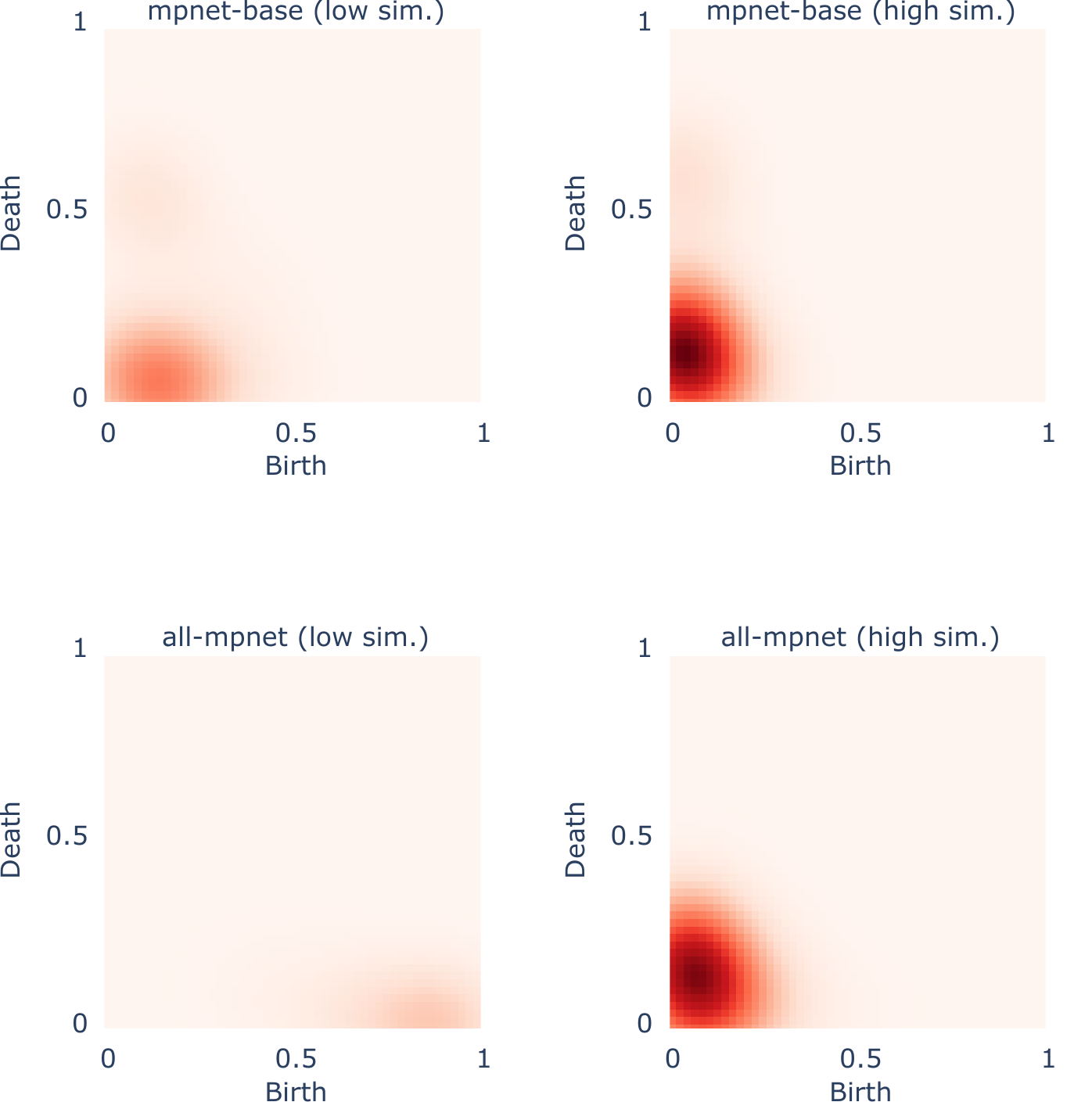}
    \captionsetup{width=0.9\columnwidth}
    \caption{
      Mean persistence images for the models mpnet-base and all-mpnet-base-v2.
    }
    \vspace{1em}
    \label{fig:mean_persistence_images_mpnet}
  \end{subfigure}
  \begin{subfigure}{\columnwidth}
    \includegraphics[width=\columnwidth]{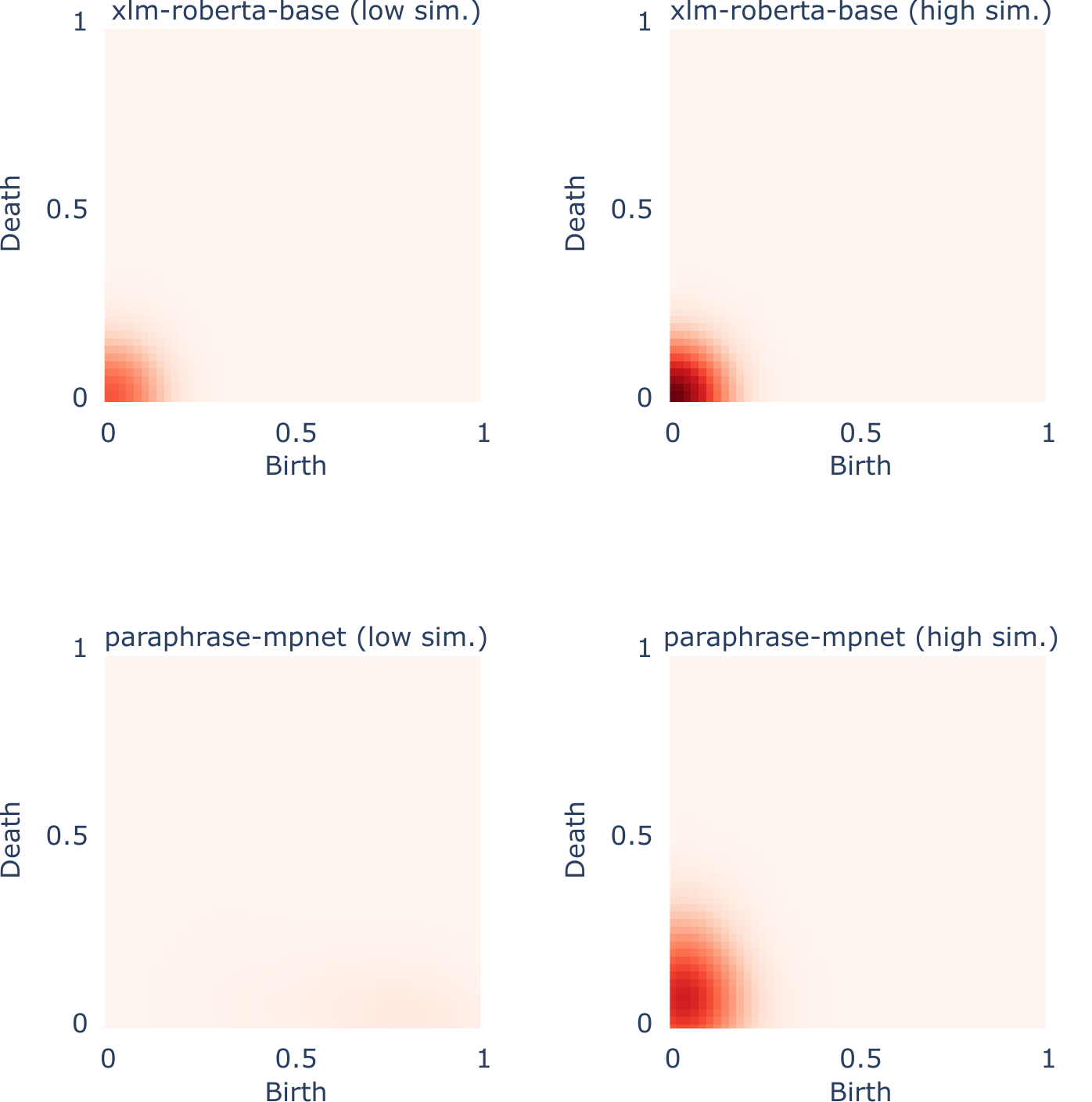}
    \captionsetup{width=0.9\columnwidth}
    \caption{
      Mean persistence images for the models xlm-roberta-base and paraphrase-multilingual-mpnet-base-v2.
    }
    \vspace{1em}
    \label{fig:mean_persistence_images_paraphrase_mpnet}
  \end{subfigure}
  \caption{
    Mean persistence images for the model pairs from Table~\ref{table:models}.
    See Figure~\ref{fig:mean_persistence_images_minilm_l6} in the main text for a description of what these images capture and how they are obtained.
  }
  \label{fig:mean_persistence_images_appendix}
\end{figure*}

See Figure~\ref{fig:mean_persistence_images_minilm_l6} in the main text and~\Cref{fig:mean_persistence_images_minilm_l12,fig:mean_persistence_images_distilroberta,fig:mean_persistence_images_mpnet,fig:mean_persistence_images_paraphrase_mpnet}.

\subsubsection{Layer Regression Experiment}\label{appendix:layer_regression_results}

\begin{figure*}[t]
  \centering
  \includegraphics[width=0.49\linewidth]{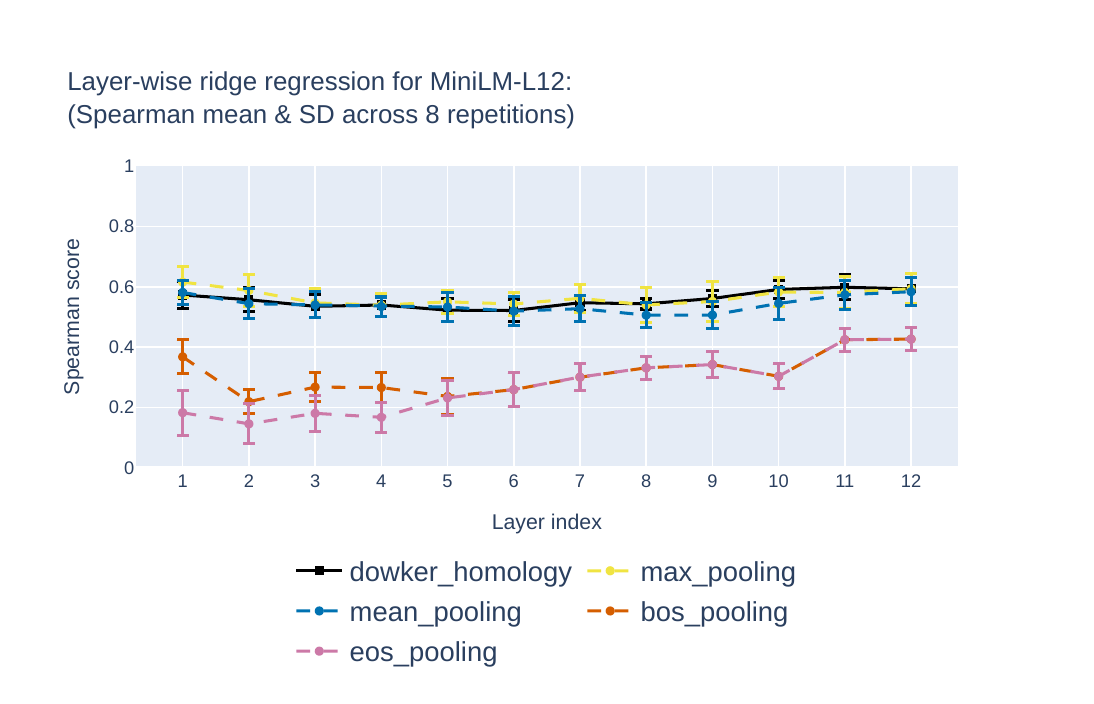}
  \hfill
  \includegraphics[width=0.49\linewidth]{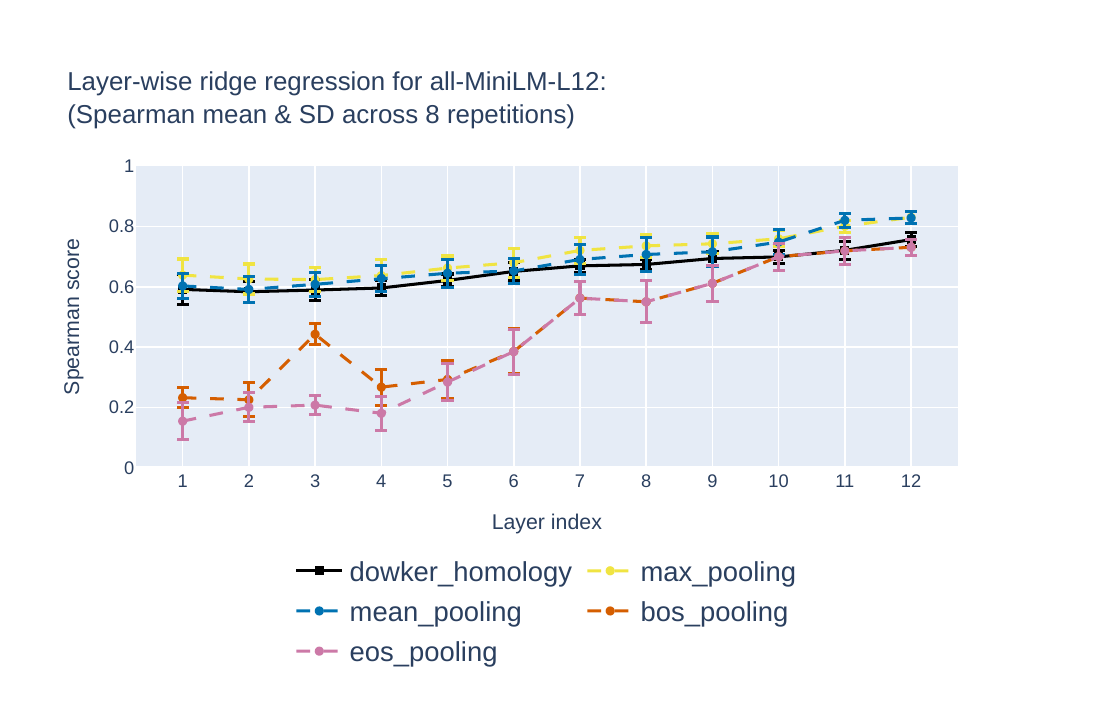}
  \caption{
    Results of the layer regression experiment for MiniLM-L12-H384-uncased (left) and all-MiniLM-L12-v2 (right).
  }
  \label{fig:layer_regression_minilm_l12}
\end{figure*}

\begin{figure*}[t]
  \centering
  \includegraphics[width=0.49\linewidth]{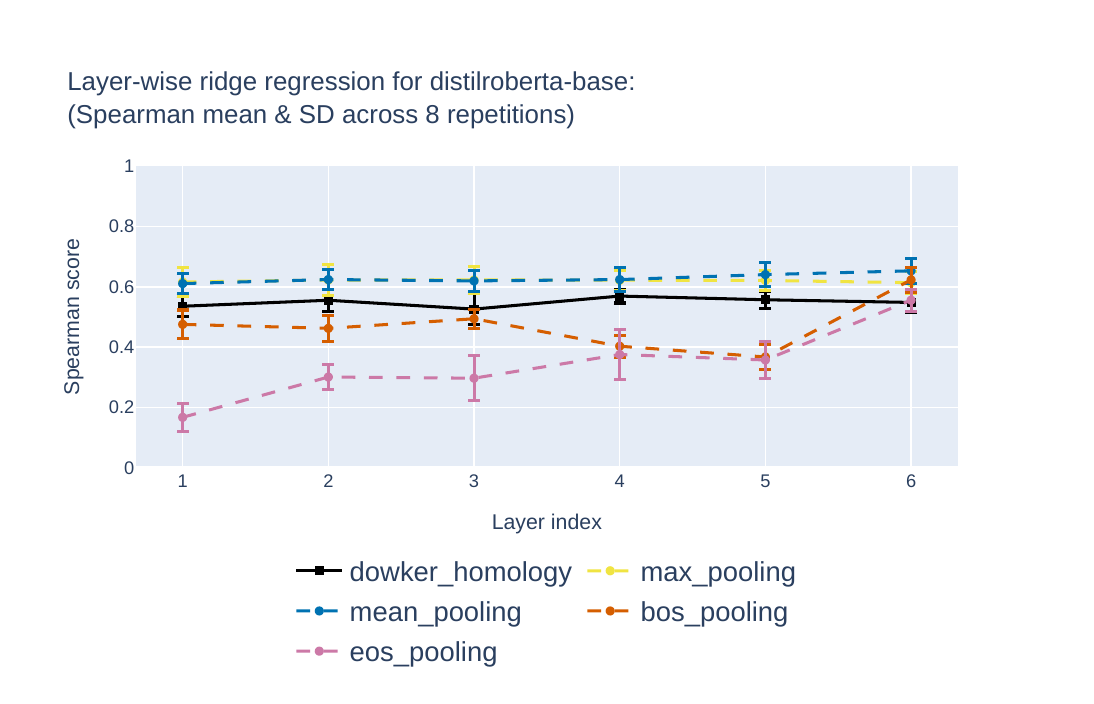}
  \hfill
  \includegraphics[width=0.49\linewidth]{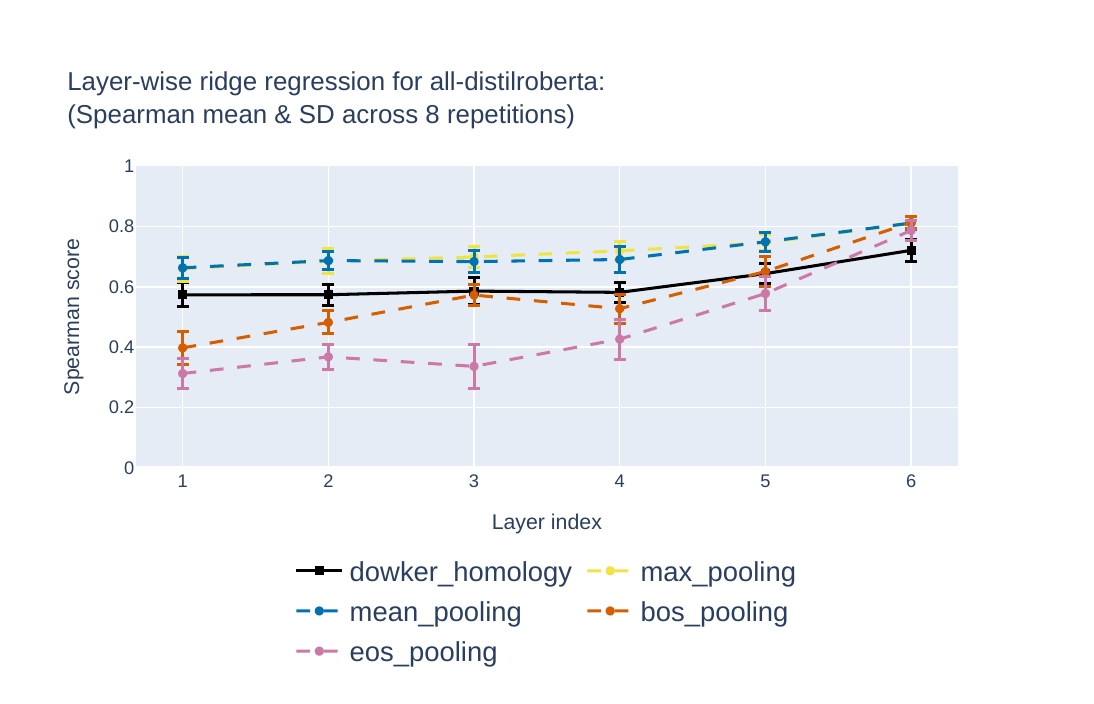}
  \caption{
    Results of the layer regression experiment for distilroberta-base (left) and all-distilroberta-v1 (right).
  }
  \label{fig:layer_regression_distilroberta}
\end{figure*}

\begin{figure*}[t]
  \centering
  \includegraphics[width=0.49\linewidth]{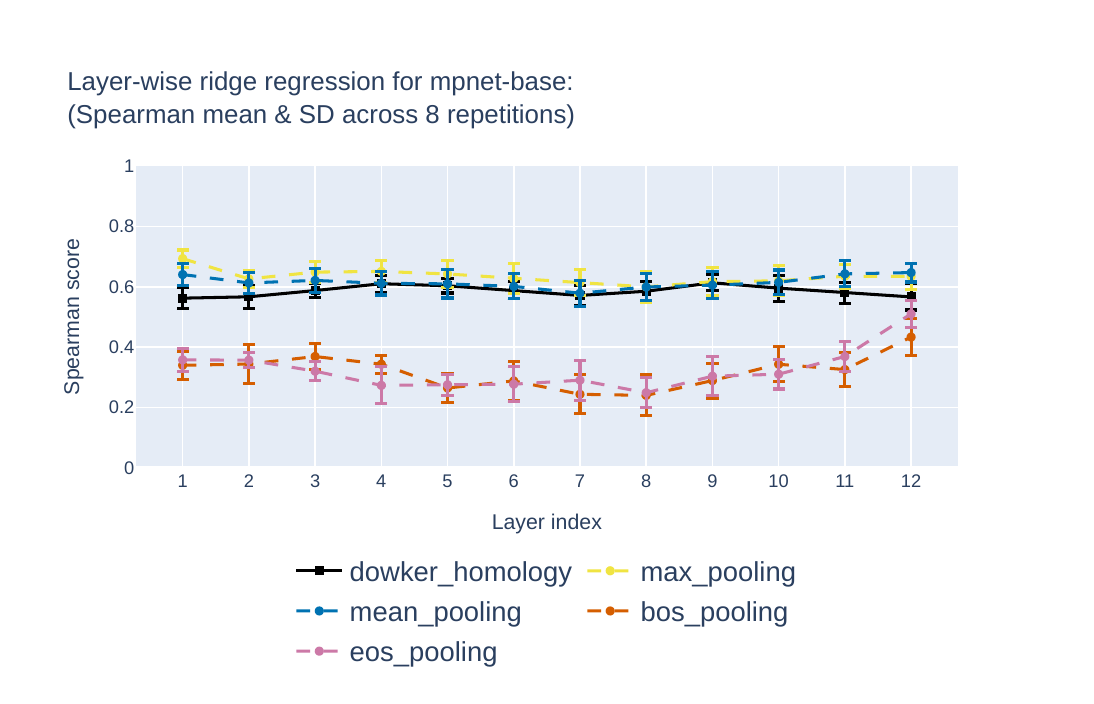}
  \hfill
  \includegraphics[width=0.49\linewidth]{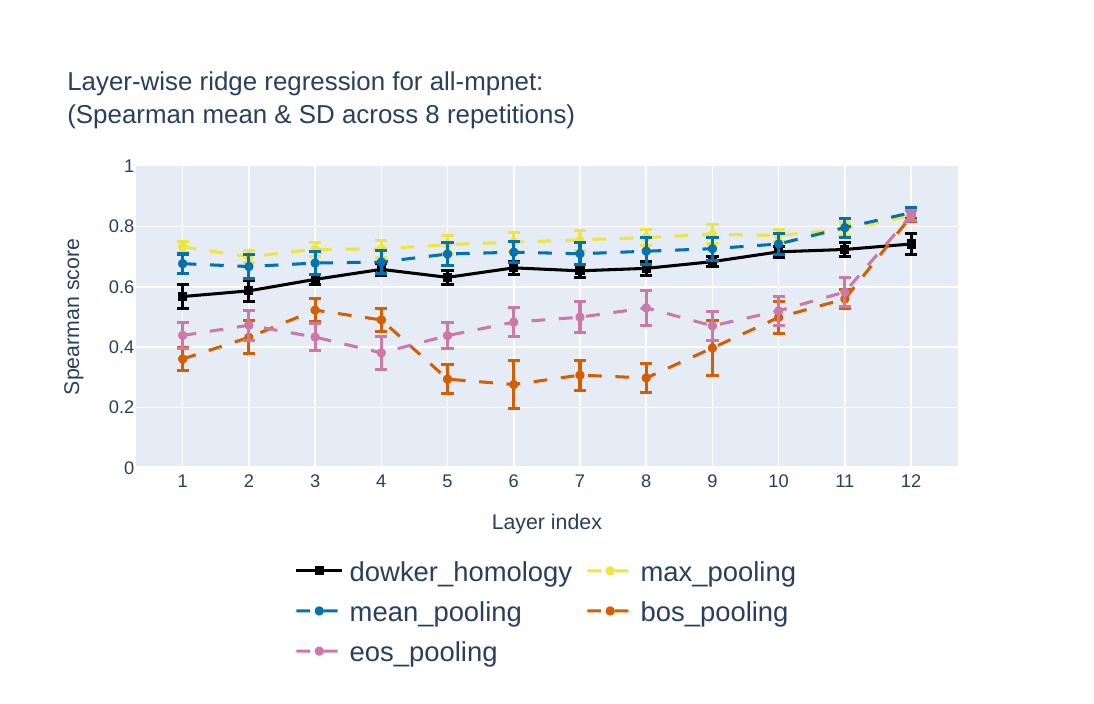}
  \caption{
    Results of the layer regression experiment for mpnet-base (left) and all-mpnet-base-v2 (right).
  }
  \label{fig:layer_regression_mpnet}
\end{figure*}

\begin{figure*}[t]
  \centering
  \includegraphics[width=0.49\linewidth]{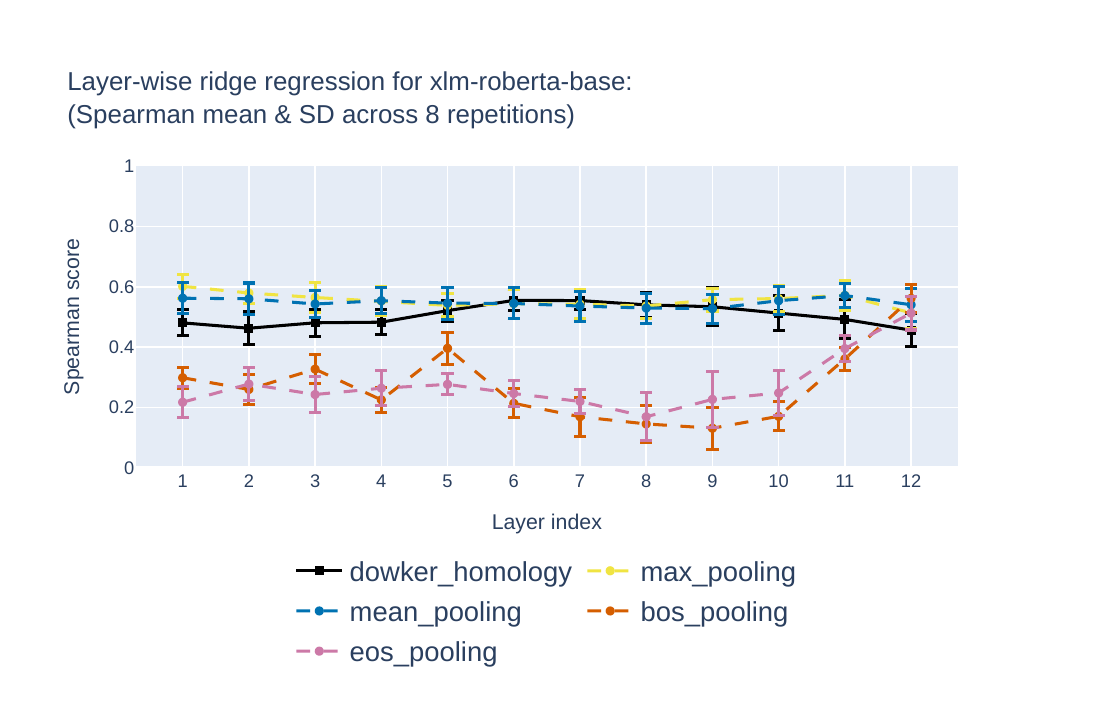}
  \hfill
  \includegraphics[width=0.49\linewidth]{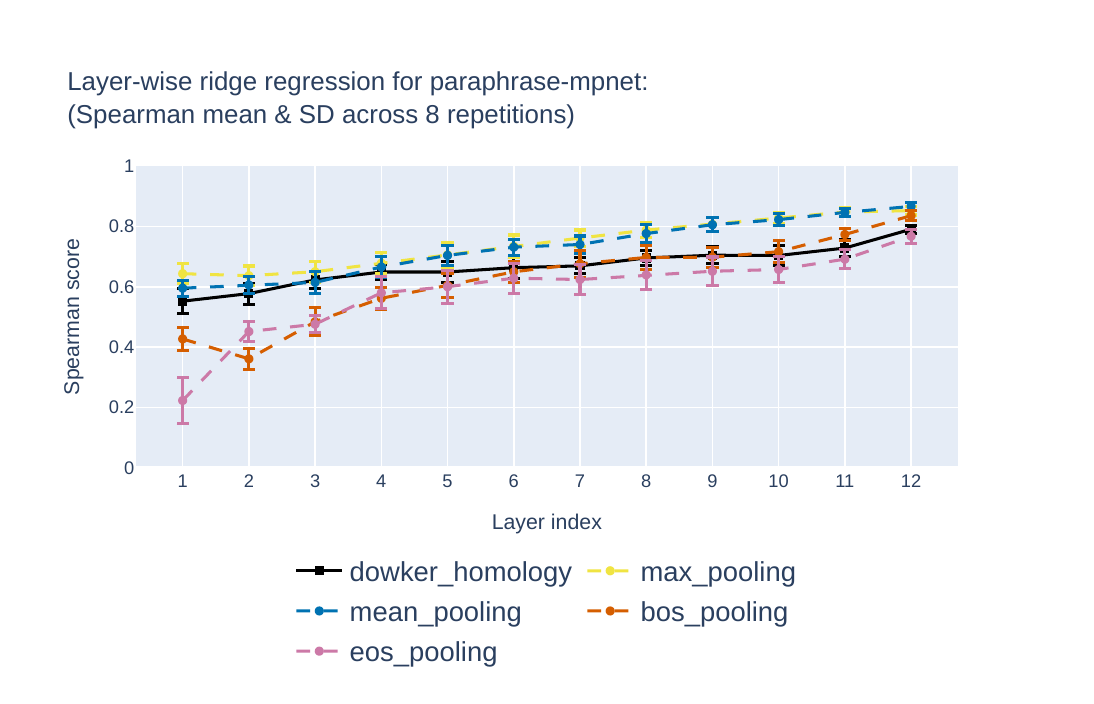}
  \caption{
    Results of the layer regression experiment for xlm-roberta-base (left) and paraphrase-multilingual-mpnet-base-v2 (right).
  }
  \label{fig:layer_regression_paraphrase_mpnet}
\end{figure*}

See Figure~\ref{fig:layer_regression_minilm_l6} in the main text and~\Cref{fig:layer_regression_minilm_l12,fig:layer_regression_distilroberta,fig:layer_regression_mpnet,fig:layer_regression_paraphrase_mpnet}.

\subsubsection{Layer Correlation Experiment}\label{appendix:layer_correlation_results}

\begin{figure*}[t]
  \centering
  \includegraphics[width=0.49\linewidth]{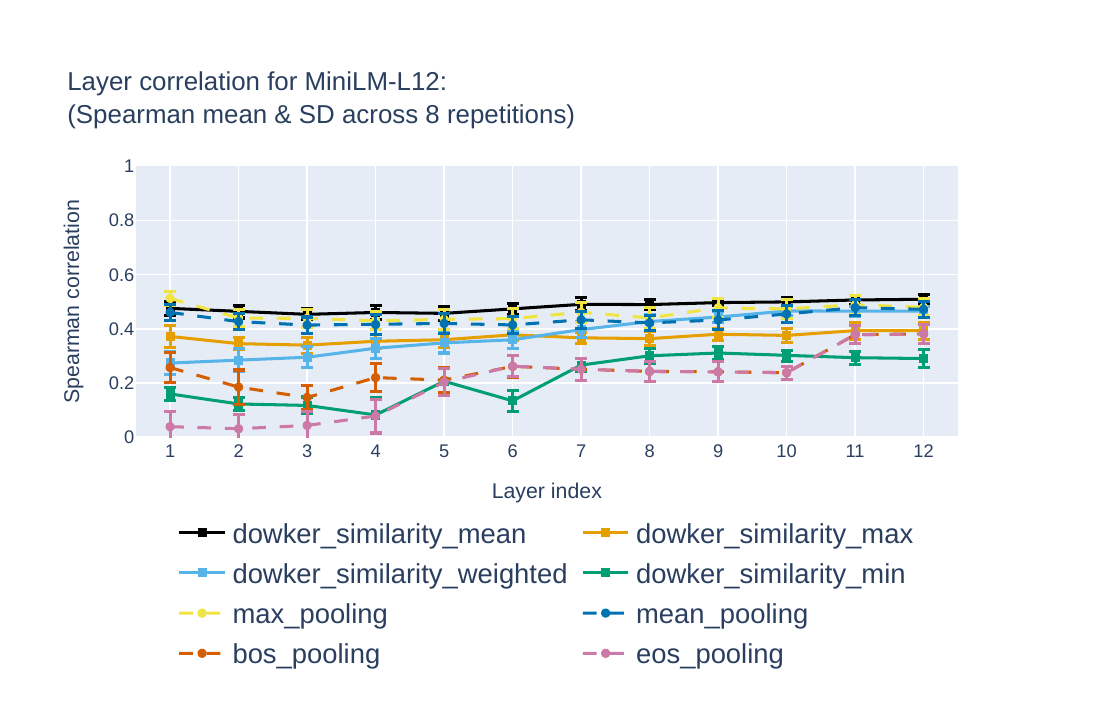}
  \hfill
  \includegraphics[width=0.49\linewidth]{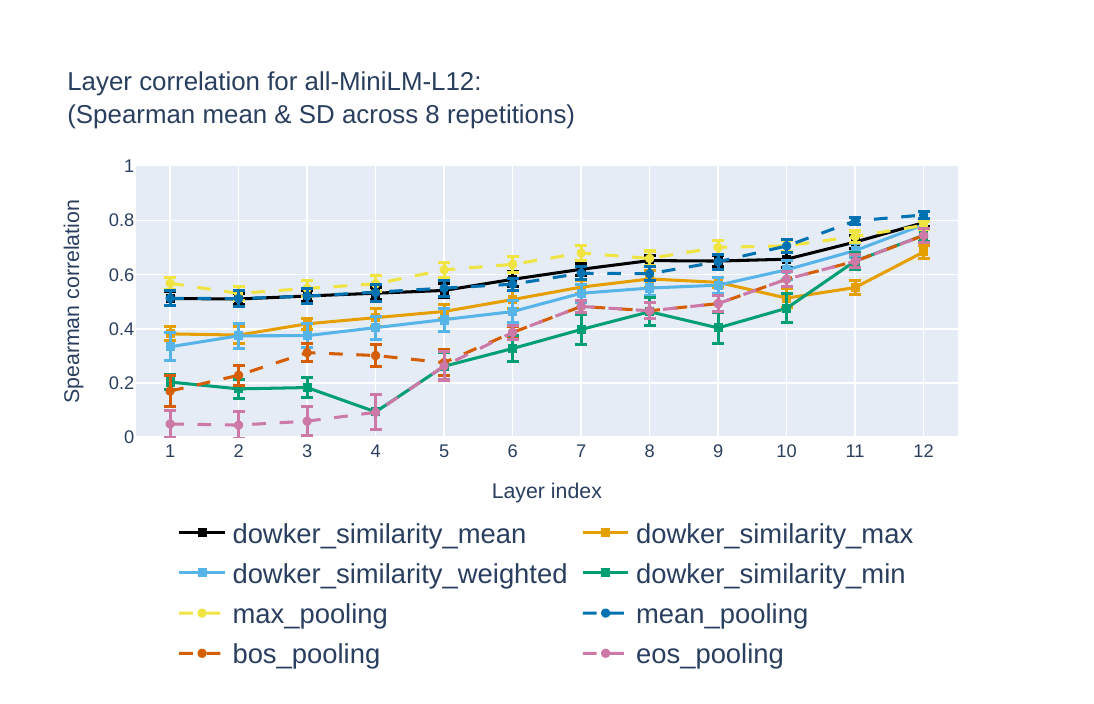}
  \caption{
    Results of the layer correlation experiment for MiniLM-L12-H384-uncased (left) and all-MiniLM-L12-v2 (right).
    Method names in the legend are sorted according to decreasing Spearman correlation averaged across model layers.
  }
  \label{fig:layer_correlation_minilm_l12}
\end{figure*}

\begin{figure*}[t]
  \centering
  \includegraphics[width=0.49\linewidth]{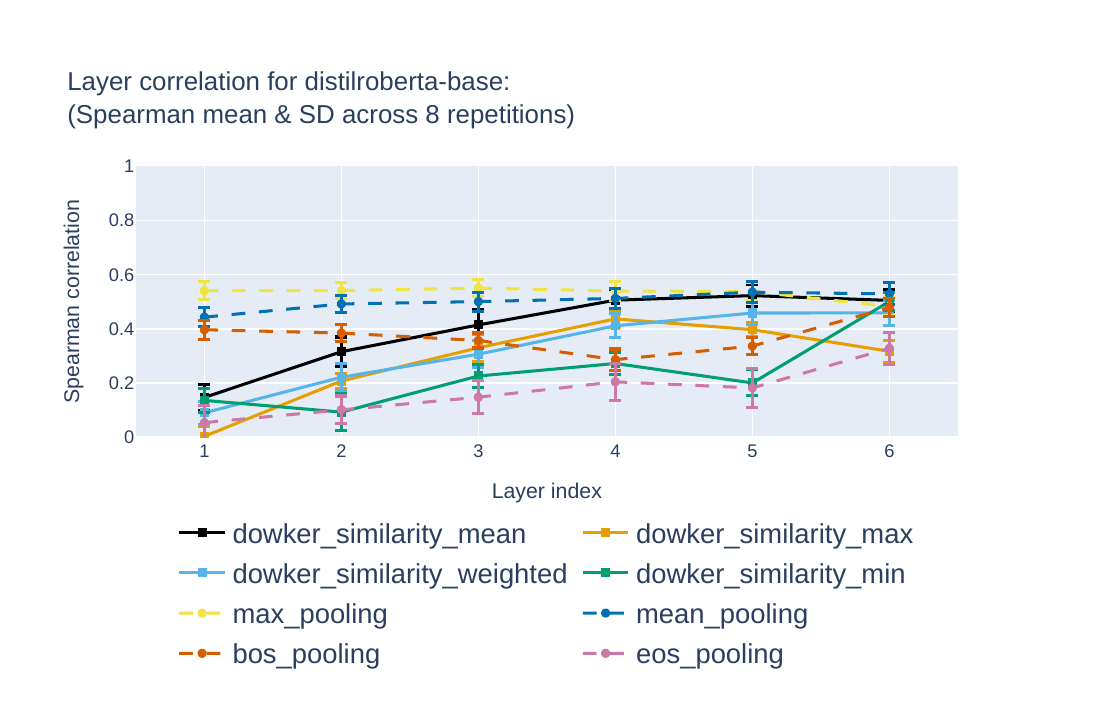}
  \hfill
  \includegraphics[width=0.49\linewidth]{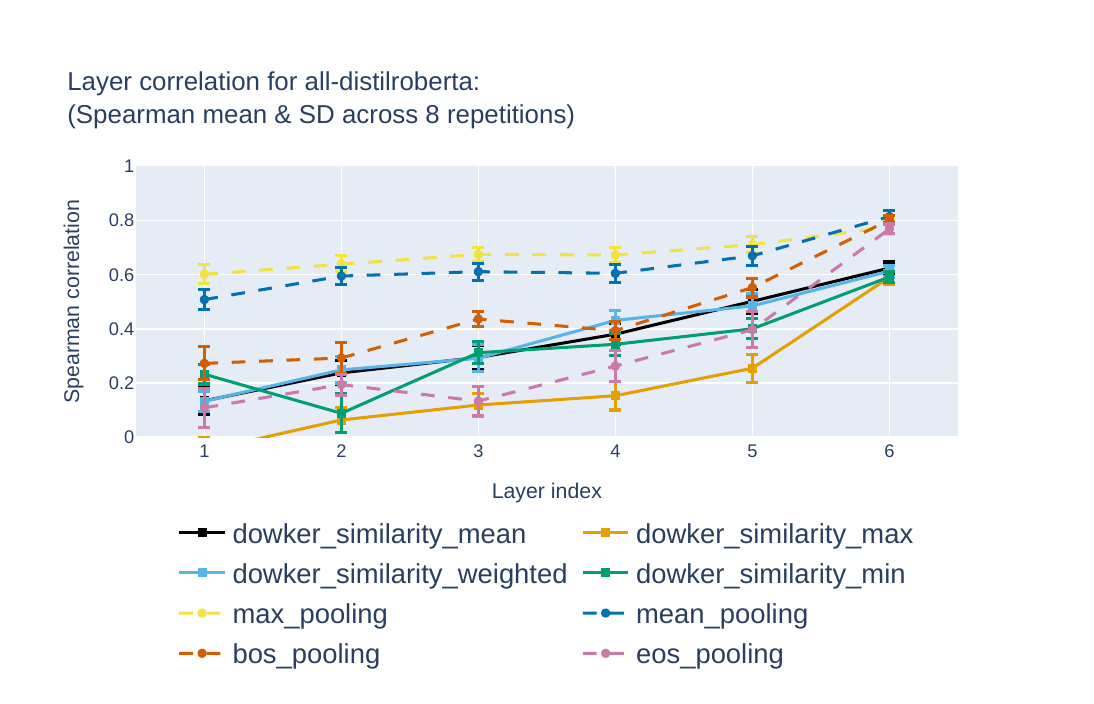}
  \caption{
    Results of the layer correlation experiment for distilroberta-base (left) and all-distilroberta-v1 (right).
    Method names in the legend are sorted according to decreasing Spearman correlation averaged across model layers.
  }
  \label{fig:layer_correlation_distilroberta}
\end{figure*}

\begin{figure*}[t]
  \centering
  \includegraphics[width=0.49\linewidth]{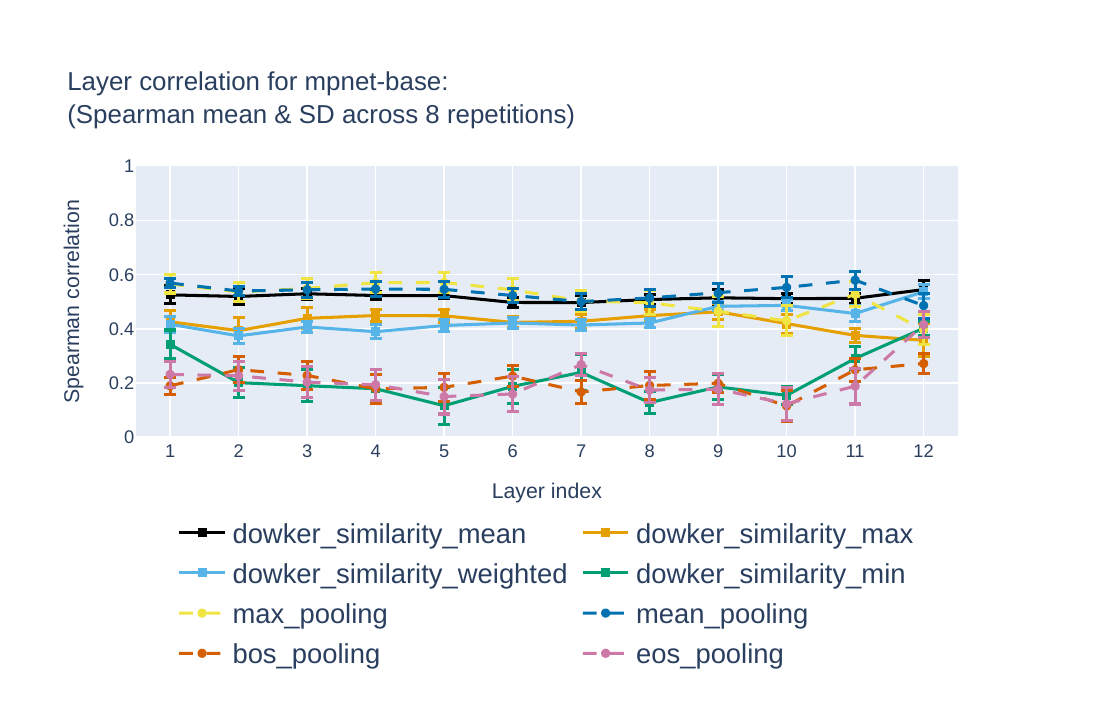}
  \hfill
  \includegraphics[width=0.49\linewidth]{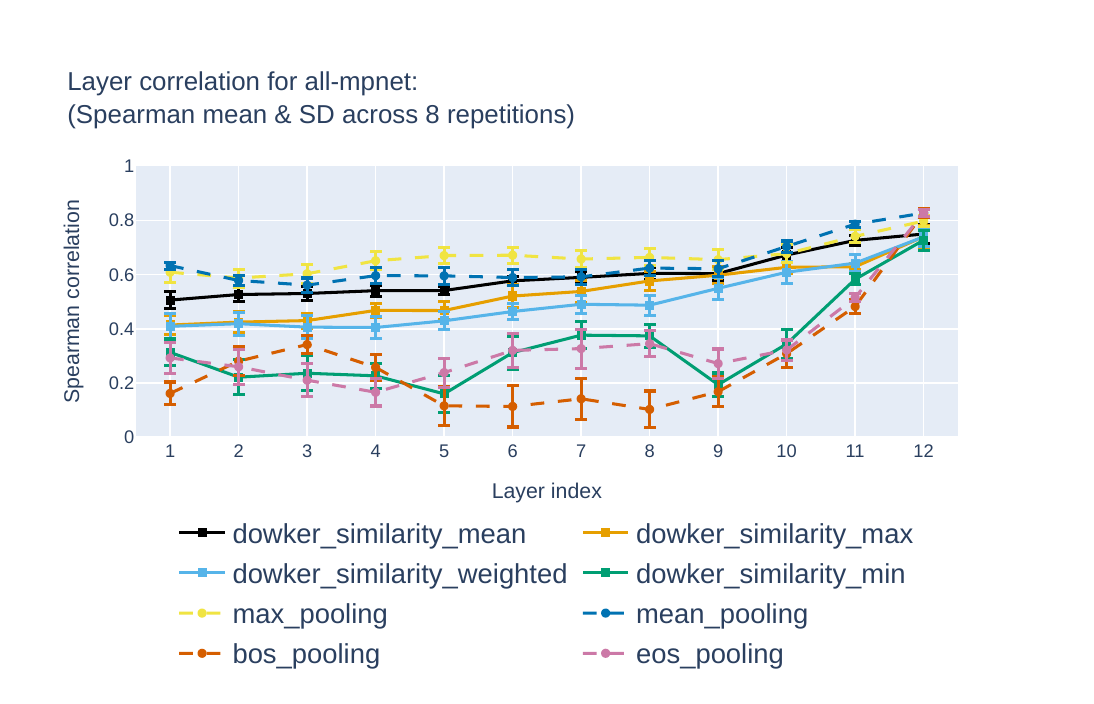}
  \caption{
    Results of the layer correlation experiment for mpnet-base (left) and all-mpnet-base-v2 (right).
    Method names in the legend are sorted according to decreasing Spearman correlation averaged across model layers.
  }
  \label{fig:layer_correlation_mpnet}
\end{figure*}

\begin{figure*}[t]
  \centering
  \includegraphics[width=0.49\linewidth]{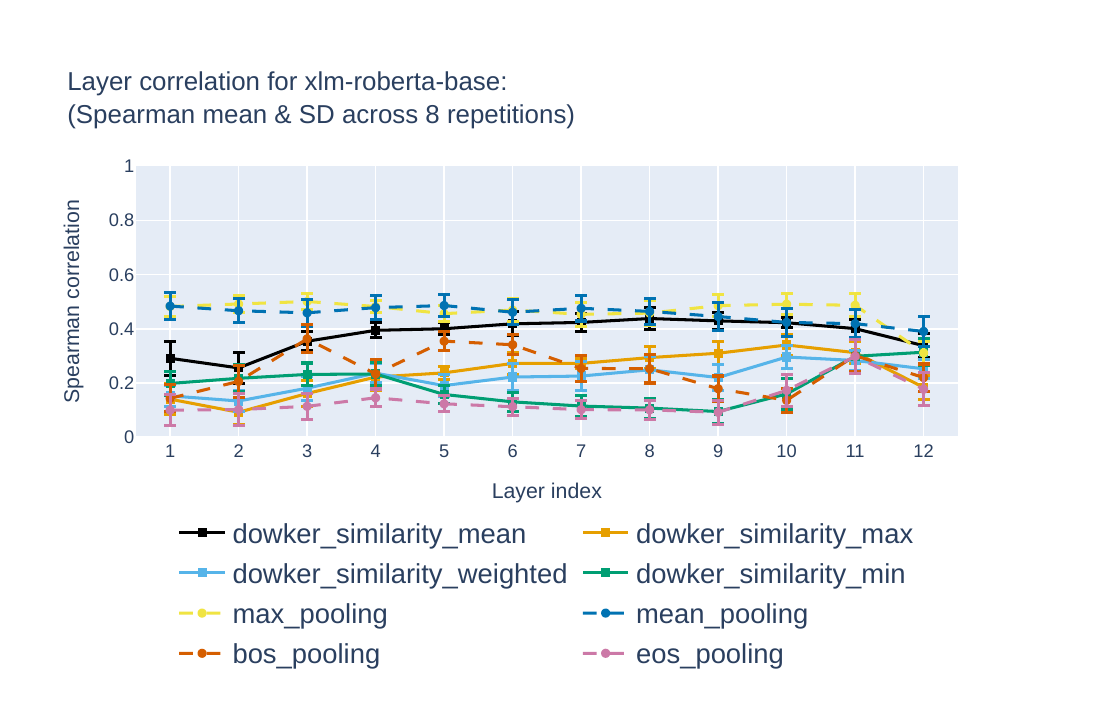}
  \hfill
  \includegraphics[width=0.49\linewidth]{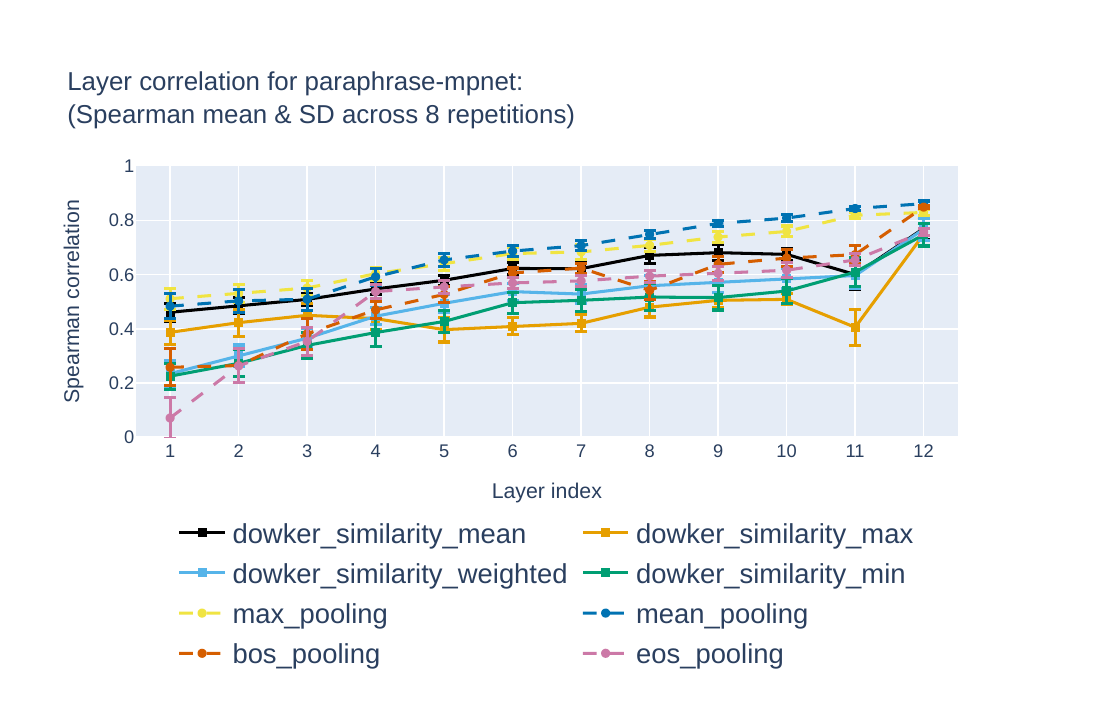}
  \caption{
    Results of the layer correlation experiment for xlm-roberta-base (left) and paraphrase-multilingual-mpnet-base-v2 (right).
    Method names in the legend are sorted according to decreasing Spearman correlation averaged across model layers.
  }
  \label{fig:layer_correlation_paraphrase_mpnet}
\end{figure*}

See Figure~\ref{fig:layer_correlation_minilm_l6} in the main text and~\Cref{fig:layer_correlation_minilm_l12,fig:layer_correlation_distilroberta,fig:layer_correlation_mpnet,fig:layer_correlation_paraphrase_mpnet}.

\end{document}

%% file: abstract.tex
\dtext homology is a topological tool that may be used to analyze the relative position of two point clouds living in a common space.
We investigate whether \dtext homology captures sentence similarity information by treating the embeddings of the tokens that constitute a sentence pair as a pair of point clouds in the latent space of a transformer model, using both models that have and have not been fine-tuned for sentence similarity.
We find that \dtext homology captures sentence similarity information, as measured by regressing \dtext homology features onto ground-truth similarity scores, and that it can be used for visual inspection of similarity data and models.
In an attempt to make \dtext homology readily applicable, we derive from it single-number summaries that we expect to capture sentence similarity directly.
These turn out to work reasonably well, but without outperforming standard sentence similarity measures based on established pooling methods.\footnote{
  All code to run experiments is available at \url{https://github.com/m-a-huber/dowker-for-sent-sim} and archived at \url{https://doi.org/10.5281/zenodo.21810253}.
}

%% file: figures/persistence_image.pdf_tex
\begingroup%
  \makeatletter%
  \providecommand\color[2][]{%
    \errmessage{(Inkscape) Color is used for the text in Inkscape, but the package 'color.sty' is not loaded}%
    \renewcommand\color[2][]{}%
  }%
  \providecommand\transparent[1]{%
    \errmessage{(Inkscape) Transparency is used (non-zero) for the text in Inkscape, but the package 'transparent.sty' is not loaded}%
    \renewcommand\transparent[1]{}%
  }%
  \providecommand\rotatebox[2]{#2}%
  \newcommand*\fsize{\dimexpr\f@size pt\relax}%
  \newcommand*\lineheight[1]{\fontsize{\fsize}{#1\fsize}\selectfont}%
  \ifx\svgwidth\undefined%
    \setlength{\unitlength}{384.26293945bp}%
    \ifx\svgscale\undefined%
      \relax%
    \else%
      \setlength{\unitlength}{\unitlength * \real{\svgscale}}%
    \fi%
  \else%
    \setlength{\unitlength}{\svgwidth}%
  \fi%
  \global\let\svgwidth\undefined%
  \global\let\svgscale\undefined%
  \makeatother%
  \begin{picture}(1,0.23959679)%
    \lineheight{1}%
    \setlength\tabcolsep{0pt}%
    \put(0,0){\includegraphics[width=\unitlength,page=1]{persistence_image.pdf}}%
    \put(0.12832563,0.00011017){\color[rgb]{0.16470588,0.24705882,0.37254902}\makebox(0,0)[t]{\lineheight{1.25}\smash{\begin{tabular}[t]{c}birth\end{tabular}}}}%
    \put(0.00543731,0.13224843){\color[rgb]{0.16470588,0.24705882,0.37254902}\rotatebox{90}{\makebox(0,0)[t]{\lineheight{1.25}\smash{\begin{tabular}[t]{c}death\end{tabular}}}}}%
    \put(0,0){\includegraphics[width=\unitlength,page=2]{persistence_image.pdf}}%
    \put(0.52100437,0.00007692){\color[rgb]{0.16470588,0.24705882,0.37254902}\makebox(0,0)[t]{\lineheight{1.25}\smash{\begin{tabular}[t]{c}birth\end{tabular}}}}%
    \put(0.39809803,0.13223274){\color[rgb]{0.16470588,0.24705882,0.37254902}\rotatebox{90}{\makebox(0,0)[t]{\lineheight{1.25}\smash{\begin{tabular}[t]{c}death--birth\end{tabular}}}}}%
    \put(0,0){\includegraphics[width=\unitlength,page=3]{persistence_image.pdf}}%
  \end{picture}%
\endgroup%

%% file: tables/table_models.tex
\begin{table*}[t]
  \centering
  \begin{tabular}{lllr}
    \hline
    \textbf{Base model}     & \textbf{Sentence transformer}         \\
    \hline
    MiniLM-L6-H384-uncased  & all-MiniLM-L6-v2                      \\
    MiniLM-L12-H384-uncased & all-MiniLM-L12-v2                     \\
    distilroberta-base      & all-distilroberta-v1                  \\
    mpnet-base              & all-mpnet-base-v2                     \\
    xlm-roberta-base        & paraphrase-multilingual-mpnet-base-v2 \\
    \hline
  \end{tabular}
  \caption{Sentence transformers and respective base models used in our experiments.}
  \label{table:models}
\end{table*}

%% file: references.bib
@article{adams_persistence_images,
  author  = {Henry Adams and Tegan Emerson and Michael Kirby and Rachel Neville and Chris Peterson and Patrick Shipman and Sofya Chepushtanova and Eric Hanson and Francis Motta and Lori Ziegelmeier},
  journal = {Journal of Machine Learning Research},
  number  = {8},
  pages   = {1--35},
  title   = {Persistence Images: A Stable Vector Representation of Persistent Homology},
  url     = {http://jmlr.org/papers/v18/16-337.html},
  volume  = {18},
  year    = {2017}
}

@inproceedings{barannikov_representation_topology_divergence,
  author    = {Barannikov, Serguei and Trofimov, Ilya and Balabin, Nikita and Burnaev, Evgeny},
  booktitle = {Proceedings of the 39th International Conference on Machine Learning},
  editor    = {Chaudhuri, Kamalika and Jegelka, Stefanie and Song, Le and Szepesvari, Csaba and Niu, Gang and Sabato, Sivan},
  month     = {17--23 Jul},
  pages     = {1607--1626},
  publisher = {PMLR},
  series    = {Proceedings of Machine Learning Research},
  title     = {Representation Topology Divergence: A Method for Comparing Neural Network Representations.},
  url       = {https://proceedings.mlr.press/v162/barannikov22a.html},
  volume    = {162},
  year      = {2022}
}

@inproceedings{cer-etal-2017-semeval,
  address   = {Vancouver, Canada},
  author    = {Cer, Daniel  and
               Diab, Mona  and
               Agirre, Eneko  and
               Lopez-Gazpio, I{\~n}igo  and
               Specia, Lucia},
  booktitle = {Proceedings of the 11th International Workshop on Semantic Evaluation ({S}em{E}val-2017)},
  doi       = {10.18653/v1/S17-2001},
  editor    = {Bethard, Steven  and
               Carpuat, Marine  and
               Apidianaki, Marianna  and
               Mohammad, Saif M.  and
               Cer, Daniel  and
               Jurgens, David},
  month     = aug,
  pages     = {1--14},
  publisher = {Association for Computational Linguistics},
  title     = {{S}em{E}val-2017 Task 1: Semantic Textual Similarity Multilingual and Crosslingual Focused Evaluation},
  url       = {https://aclanthology.org/S17-2001/},
  year      = {2017}
}

@article{chowdhury_memoli_functorial_dowker_theorem,
  author   = {Chowdhury, Samir and M\'emoli, Facundo},
  doi      = {10.1007/s41468-018-0020-6},
  fjournal = {Journal of Applied and Computational Topology},
  issn     = {2367-1726,2367-1734},
  journal  = {J. Appl. Comput. Topol.},
  mrclass  = {68U05 (55N35)},
  mrnumber = {3873182},
  number   = {1-2},
  pages    = {115--175},
  title    = {A functorial {D}owker theorem and persistent homology of asymmetric networks},
  url      = {https://doi.org/10.1007/s41468-018-0020-6},
  volume   = {2},
  year     = {2018}
}

@article{dowker_homology_groups_of_relations,
  author     = {Dowker, C. H.},
  doi        = {10.2307/1969768},
  fjournal   = {Annals of Mathematics. Second Series},
  issn       = {0003-486X},
  journal    = {Ann. of Math. (2)},
  mrclass    = {56.0X},
  mrnumber   = {48030},
  mrreviewer = {E.\ H.\ Spanier},
  pages      = {84--95},
  title      = {Homology groups of relations},
  url        = {https://doi.org/10.2307/1969768},
  volume     = {56},
  year       = {1952}
}

@incollection{gudhi:PersistenceRepresentations,
  author    = {Pawel Dlotko},
  booktitle = {GUDHI User and Reference Manual},
  edition   = {3.12.0},
  publisher = {GUDHI Editorial Board},
  title     = {Persistence representations},
  url       = {https://gudhi.inria.fr/doc/3.12.0/group___persistence__representations.html},
  year      = {2026}
}

@inproceedings{huber_2026_fixation_sequences,
  address   = {New York, NY, USA},
  articleno = {12},
  author    = {Huber, Marius and Reich, David Robert and J\"{a}ger, Lena Ann},
  booktitle = {Proceedings of the 2026 Symposium on Eye Tracking Research and Applications},
  doi       = {10.1145/3797246.3803045},
  isbn      = {9798400725197},
  location  = {
               },
  numpages  = {13},
  publisher = {Association for Computing Machinery},
  series    = {ETRA '26},
  title     = {Fixation Sequences as Time Series: A Topological Approach to Dyslexia Detection},
  url       = {https://doi.org/10.1145/3797246.3803045},
  year      = {2026}
}

@misc{huber_schnider_flagifying_dowker_complex,
  archiveprefix = {arXiv},
  author        = {Marius Huber and Patrick Schnider},
  eprint        = {2508.08025},
  note          = {arXiv:2508.08025 [math.AT], \url{https://arxiv.org/abs/2508.08025}},
  primaryclass  = {math.AT},
  title         = {Flagifying the Dowker Complex},
  url           = {https://arxiv.org/abs/2508.08025},
  year          = {2025}
}

@inproceedings{michel_geometry_of_word_embeddings,
  address   = {Vancouver, Canada},
  author    = {Michel, Paul  and
               Ravichander, Abhilasha  and
               Rijhwani, Shruti},
  booktitle = {Proceedings of the 2nd Workshop on Representation Learning for {NLP}},
  doi       = {10.18653/v1/W17-2628},
  editor    = {Blunsom, Phil  and
               Bordes, Antoine  and
               Cho, Kyunghyun  and
               Cohen, Shay  and
               Dyer, Chris  and
               Grefenstette, Edward  and
               Hermann, Karl Moritz  and
               Rimell, Laura  and
               Weston, Jason  and
               Yih, Scott},
  month     = aug,
  pages     = {235--240},
  publisher = {Association for Computational Linguistics},
  title     = {Does the Geometry of Word Embeddings Help Document Classification? {A} Case Study on Persistent Homology-Based Representations},
  url       = {https://aclanthology.org/W17-2628/},
  year      = {2017}
}

@inproceedings{reimers-gurevych-2019-sentence,
  address   = {Hong Kong, China},
  author    = {Reimers, Nils  and
               Gurevych, Iryna},
  booktitle = {Proceedings of the 2019 Conference on Empirical Methods in Natural Language Processing and the 9th International Joint Conference on Natural Language Processing (EMNLP-IJCNLP)},
  doi       = {10.18653/v1/D19-1410},
  editor    = {Inui, Kentaro  and
               Jiang, Jing  and
               Ng, Vincent  and
               Wan, Xiaojun},
  month     = nov,
  pages     = {3982--3992},
  publisher = {Association for Computational Linguistics},
  title     = {Sentence-{BERT}: Sentence Embeddings using {S}iamese {BERT}-Networks},
  url       = {https://aclanthology.org/D19-1410/},
  year      = {2019}
}

@article{scikit-learn,
  author  = {Pedregosa, F. and Varoquaux, G. and Gramfort, A. and Michel, V.
             and Thirion, B. and Grisel, O. and Blondel, M. and Prettenhofer, P.
             and Weiss, R. and Dubourg, V. and Vanderplas, J. and Passos, A. and
             Cournapeau, D. and Brucher, M. and Perrot, M. and Duchesnay, E.},
  journal = {Journal of Machine Learning Research},
  pages   = {2825--2830},
  title   = {Scikit-learn: Machine Learning in {P}ython},
  volume  = {12},
  year    = {2011}
}
